\documentclass[12pt,a4paper,oneside]{article}
\usepackage[T1]{fontenc}
\usepackage[utf8]{inputenc}
\usepackage{amssymb,amsfonts,mathtools,amsthm}
\usepackage{graphicx}
\usepackage{placeins}
\usepackage{caption}
\graphicspath{{./}}
\usepackage{booktabs}
\usepackage{tabularx}
\usepackage{enumitem}
\usepackage{xcolor}

\newif\ifshowrevisions
\showrevisionstrue

\usepackage[sectionbib,authoryear,round]{natbib}
\usepackage{hyperref}
\usepackage{xurl}
\usepackage{setspace}
\usepackage[nameinlink]{cleveref}
\hypersetup{hidelinks}
\setlist[itemize]{leftmargin=*,itemsep=2pt,topsep=4pt}
\setlist[enumerate]{leftmargin=*,itemsep=2pt,topsep=4pt}
\numberwithin{equation}{section}
\newtheorem{theorem}{Theorem}[section]

\newtheorem{proposition}[theorem]{Proposition}
\newtheorem{corollary}[theorem]{Corollary}
\newtheorem{assumption}[theorem]{Assumption}
\theoremstyle{definition}
\newtheorem{definition}[theorem]{Definition}
\theoremstyle{remark}

\crefname{theorem}{theorem}{theorems}
\crefname{lemma}{lemma}{lemmas}
\crefname{proposition}{proposition}{propositions}
\crefname{corollary}{corollary}{corollaries}
\crefname{assumption}{assumption}{assumptions}
\crefname{definition}{definition}{definitions}
\crefname{remark}{remark}{remarks}
\Crefname{theorem}{Theorem}{Theorems}
\Crefname{lemma}{Lemma}{Lemmas}
\Crefname{proposition}{Proposition}{Propositions}
\Crefname{corollary}{Corollary}{Corollaries}
\Crefname{assumption}{Assumption}{Assumptions}
\Crefname{definition}{Definition}{Definitions}
\Crefname{remark}{Remark}{Remarks}

\newcommand{\M}{\mathcal M}
\newcommand{\E}{\mathbb E}
\newcommand{\Pbb}{\mathbb P}
\newcommand{\R}{\mathbb R}
\newcommand{\cN}{\mathcal N}
\newcommand{\cP}{\mathcal P}
\newcommand{\grad}{\operatorname{grad}}
\newcommand{\Hess}{\operatorname{Hess}}
\newcommand{\Exp}{\operatorname{Exp}}
\newcommand{\Log}{\operatorname{Log}}
\newcommand{\argmin}{\operatorname*{arg\,min}}
\newcommand{\op}{\mathrm{op}}
\newcommand{\tr}{\operatorname{tr}}

\newcommand{\diam}{\operatorname{diam}}
\newcommand{\norm}[1]{\left\lVert #1\right\rVert}
\newcommand{\abs}[1]{\left\lvert #1\right\rvert}
\newcommand{\inner}[2]{\left\langle #1,#2\right\rangle}

\newcommand{\eps}{\varepsilon}
\newcommand{\wh}{\widehat}
\newcommand{\wt}{\widetilde}
\newcommand{\ot}{\otimes}
\newcommand{\Torus}{\mathbb T}
\newcommand{\cJ}{\mathcal J}

\newcommand{\Cov}{\operatorname{Cov}}
\usepackage{lmodern,microtype,geometry}

\hypersetup{hidelinks,
  pdftitle={Riemannian Sample-Average Approximation under Markov Dependence: Finite-Sample Guarantees, One-Step Refinement, and Inference},
  pdfauthor={Xiaotian Chang, Qirui Hu and Nan Qiao}}
\title{Riemannian Sample-Average Approximation under Markov Dependence:
Finite-Sample Guarantees, One-Step Refinement, and Inference}
\author{%
  Xiaotian Chang\thanks{Division of Mathematical Sciences,
    Nanyang Technological University, Singapore.
    E-mail: \texttt{xiaotian.chang@ntu.edu.sg}.
    },\quad
  Qirui Hu\thanks{ School of Statistics and Data Science,
    Shanghai University of Finance and Economics, China. 
    Department of Mathematics, Ruhr University Bochum. 
    Corresponding author.
    E-mail: \texttt{huqirui@mail.shufe.edu.cn}.}\quad
  and Nan Qiao\thanks{School of Mathematics, Statistics and Mechanics,
    Beijing University of Technology, China.
    E-mail: \texttt{qiaonan0915@ruc.edu.cn}. 
    Authors contributed equally to this work, and their names are listed alphabetically.}%
}
\date{}

\begin{document}\maketitle
\begin{abstract}
We study estimation and inference for an interior population-loss minimizer on a compact Riemannian manifold from a single stationary Markov trajectory. In the sample-average approximation (SAA) framework,  finite-sample bounds distinguish sampling error from numerical error under mild conditions. For exact SAA, we obtain expectation expansions for parameter and excess risk and a limiting distribution. On every compact manifold, exact SAA attains the minimax squared-error order on a constructed smooth model class that remains fixed across sample sizes. We give sufficient numerical accuracy conditions for preserving the exact-SAA limiting distribution; one intrinsic Newton correction permits coarser localized initial estimates. For geometrically ergodic Harris chains with positive definite long-run score covariance, batch means at the computed solution yield asymptotically valid geodesic confidence regions in fixed dimension. Simulations and an application to MetroPT-3 train-compressor sensor data examine estimation error, numerical accuracy, and coverage.
\end{abstract}

\setstretch{1.5}
\section{Introduction}
\label{sec:introduction}

For manifold-valued parameters, ignoring geometry distorts even basic summaries: wind directions of $359^\circ$ and $1^\circ$ both point nearly due north, yet their arithmetic average points due south whereas their circular mean on the unit circle points north.
Similarly, in robotics, averaging the entries of two rotation matrices can yield a matrix that represents no possible orientation. Geometry also matters when the observations are ordinary vectors. In the MetroPT-3 urban train air-compressor data \citep{davari2021metropt}, the leading subspace of pressure, temperature, and current measurements summarizes how these signals vary together. Because different orthonormal bases span the same subspace, this operating summary is naturally a point on a Grassmann manifold.

\begin{figure}[!htb]
\centering
\includegraphics[width=\linewidth]{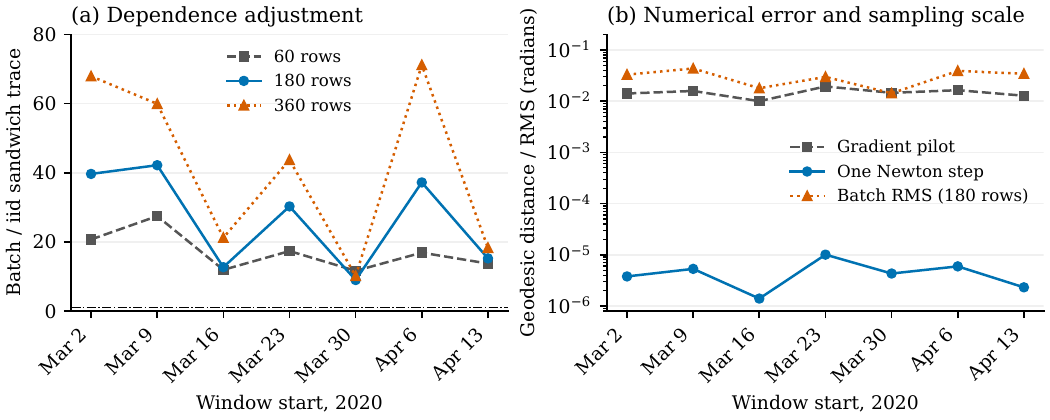}
\caption{Seven MetroPT-3 pre-fault windows. (a) Batch-to-i.i.d. ratio of estimated total subspace variance. (b) Pilot and one-step distances to the exact empirical subspace, alongside root-mean-square sampling uncertainty estimated from 180-reading batches.}
\label{fig:metropt}
\end{figure}

Consecutive compressor readings are serially correlated, and the operating subspace is estimated from a single stream of observations. More generally, many manifold summaries are defined by minimizing a population loss over a compact Riemannian manifold. Under stationary Markov sampling, sample-average approximation (SAA) replaces the population loss with its average along a single trajectory, and an iterative algorithm returns an approximate minimizer, introducing numerical optimization error alongside sampling variability. Assessing the accuracy and uncertainty of the resulting estimate therefore requires accounting for both temporal dependence and solver inaccuracy. Figure~\ref{fig:metropt} illustrates both effects in the MetroPT-3 data: accounting for serial dependence substantially increases estimated subspace uncertainty, while one Newton step removes nearly all numerical error from a coarsely stopped gradient iterate.

Classical SAA theory relates uniform approximation of the objective to stability and asymptotic distributions of its minimizers \citep{shapiro1993,shapirohomemdemello2000,shapirodentchevaruszczynski2021}. For non-i.i.d.\ sampling, \citet{homemdemello2008} transfers pointwise convergence rates to SAA optimizers, covering Latin hypercube and randomized quasi-Monte Carlo designs. For serially dependent observations, \citet{wang2022saa} give guarantees under summable $\phi$-mixing, and \citet{wang2024conditional} treat dependent conditional stochastic optimization. For random Euclidean objectives, \citet{bassettderide2022} establish one-step equivalence for scaled proximal and proximal-gradient estimators. Riemannian optimization provides algorithms for smooth constrained problems \citep{absil2008,bonnabel2013,tripuraneni2018,boumal2023}, including stochastic approximation driven by Markov data \citep{durmus2021,karimi2022}.

For independent manifold-valued observations, early work developed limit theory for Fr\'echet means and methods for intrinsic inference \citep{bhattacharya2003,bhattacharya2005,pennec2006,bhattacharyalin2017}. Later studies address cut-locus effects, robust estimation, and degenerate Hessians \citep{hotzlewood2024,leejung2026,leejunghuber2026,eltznerhuckemann2019}. Related results include finite-sample risk bounds \citep{ahidarcoutrix2020,brunel2023,brunelserres2024,escande2024}, lower bounds near nonunique Fr\'echet means \citep{hundrieser2026}, and empirical likelihood for intrinsic inference \citep{kurisuotsu2026}. For manifold time series, ergodic Fr\'echet-mean theory gives consistency \citep{abuqraispigoli2026}, and local score central limit theorems yield distributional limits for Markov Fr\'echet means \citep{abuqrais2026clt}. Long-run covariance estimation from one trajectory is well developed in Markov chain Monte Carlo \citep{flegaljones2010,vatsflegaljones2019}. However, existing results treat serial dependence on manifolds and approximate optimization separately. A central question here is how accurately a general smooth manifold loss must be minimized before the computed solution and derivatives evaluated at it support valid estimation and inference from the same dependent trajectory. Unlike the exact Fr\'echet-mean theory of \citet{abuqrais2026clt} and the Euclidean one-step analysis of \citet{bassettderide2022}, our setting requires global localization along the dependent trajectory together with parallel-transport control on random, numerically perturbed tangent spaces.

Under Markov spectral contraction, our finite-sample bounds separate sampling fluctuations along the trajectory from the numerical error of an approximate minimizer.
The empirical loss can have stationary points far from the population minimizer, so a small gradient alone cannot locate the solution. We bound how much the empirical loss can fluctuate along the trajectory to determine when an approximate minimum lies near the population target. Near the population minimizer, the loss increases at least quadratically with geodesic distance in every direction. This local growth lets us bound the distance to the target in terms of sampling fluctuations and the solver's objective gap or gradient residual. 

For exact SAA, we derive limiting distributions and second-order expectation expansions for both squared geodesic error and excess population loss. If the population loss increases slowly near its minimizer in a given direction, an estimation error in that direction can contribute substantially to squared distance but little to excess loss. On any compact Riemannian manifold, we also construct a smooth loss and a Markov model family, fixed across sample sizes, on which exact SAA attains the minimax squared geodesic risk. This rate exhibits a transition from constant-order error on short trajectories to decay proportional to dimension divided by the dependence-adjusted sample size once the latter exceeds the dimension.  

The computed solution has the same first-order sampling distribution as exact SAA if its optimization error is negligible relative to sampling fluctuations. We give sufficient numerical tolerances for this equivalence and show that one Newton correction on the manifold allows a coarser localized pilot estimate near the population minimizer. In Figure~\ref{fig:metropt}b, the distance from the initial estimate to the exact empirical subspace is comparable to the estimated sampling uncertainty. One Newton step reduces that distance to less than one thousandth of the estimated uncertainty in every window.

For inference from the same trajectory, we combine the empirical Hessian with a nonoverlapping batch-means estimator of the long-run score covariance, both evaluated at the computed solution, to form geodesic Wald confidence regions. With the batch lengths specified below, these regions attain nominal coverage asymptotically in fixed dimension when the stationary chain is Harris recurrent and geometrically ergodic and the long-run score covariance is positive definite. Figure~\ref{fig:metropt}a shows that treating successive compressor readings as independent understates subspace uncertainty in all seven windows.


The rest of the paper is organized as follows. Section~\ref{sec:setup} introduces the statistical setting, computed estimators, and confidence regions. Section~\ref{sec:theory} establishes finite-sample error bounds, risk expansions, minimax rates, one-step equivalence, and coverage guarantees. Section~\ref{sec:numerical-additional} evaluates these results in simulations, Section~\ref{sec:metropt} analyzes compressor operating subspaces in MetroPT-3, and Section~\ref{sec:discussion} concludes with open directions. Appendix contains some theoretical details, simulation designs and additional numerical results.


\section{Methodology}
\label{sec:setup}
\subsection{Statistical setting}\label{sec:statistical-setting}

Let $(\M,g)$ be an $m$-dimensional Riemannian manifold with intrinsic distance $d$, measured by lengths of curves in $\M$. The observations $\xi_1,\ldots,\xi_N$ take values in a standard Borel space $(\Xi,\mathcal A)$ and have a common marginal distribution $\pi$. For a loss $\ell:\M\times\Xi\to\R$, define the population objective $f(x)=\E_\pi[\ell(x,\xi)]$ and its empirical counterpart $L_N(x)=N^{-1}\sum_{i=1}^N\ell(x,\xi_i)$. An exact SAA estimator $\wh x_N$ is a measurable global minimizer of $L_N$ and provides the benchmark for the computed estimators. We first impose global regularity conditions on the manifold and the loss function. 

\begin{assumption}
\label{ass:global}
Let $(\M,g)$ be a smooth, compact, connected, $m$-dimensional Riemannian manifold. The loss $\ell(x,z)$ is measurable in $z$ and continuous in $x$. There are constants $B,G<\infty$ such that $\abs{\ell(x,z)}\le B$ and $\abs{\ell(x,z)-\ell(y,z)}\le Gd(x,y)$ for all $z\in\Xi$ and all $x,y\in\M$.
\end{assumption}

Compactness ensures that $f$ and $L_N$ attain their minima. Let $\mathsf P$ be a Markov kernel on $\Xi$ with invariant distribution $\pi$. Its associated operator on $L^2(\pi)$ is $(\mathsf Ph)(z)=\int_\Xi h(z')\,\mathsf P(z,\mathrm{d}z')$, and $\Pi h=\int_\Xi h\,\mathrm{d}\pi$ denotes the projection onto constant functions.

\begin{assumption}
\label{ass:global}
Let $(\M,g)$ be a smooth, compact, connected, $m$-dimensional Riemannian manifold. The loss $\ell(x,z)$ is measurable in $z$ and continuous in $x$. There are constants $B,G<\infty$ such that $\abs{\ell(x,z)}\le B$ and $\abs{\ell(x,z)-\ell(y,z)}\le Gd(x,y)$ for all $z\in\Xi$ and all $x,y\in\M$.
\end{assumption}

Compactness ensures that $f$ and $L_N$ attain their minima. Let $\mathsf P$ be a Markov kernel on $\Xi$ with invariant distribution $\pi$. Its associated operator on $L^2(\pi)$ is $(\mathsf Ph)(z)=\int_\Xi h(z')\,\mathsf P(z,\mathrm{d}z')$, and $\Pi h=\int_\Xi h\,\mathrm{d}\pi$ denotes the projection onto constant functions.

\begin{assumption}
\label{ass:spectral}
The observations $(\xi_i)_{i\ge1}$ form a stationary Markov chain with transition kernel $\mathsf P$ and invariant distribution $\pi$. On the mean-zero subspace $L^2_0(\pi)=\{h\in L^2(\pi):\Pi h=0\}$, {$\lambda:=\norm{\mathsf P}_{L_0^2(\pi)\to L_0^2(\pi)}=\norm{\mathsf P-\Pi}_{L^2(\pi)\to L^2(\pi)}<1$}. Define the dependence multiplier $\tau_\lambda:=(1+\lambda)/(1-\lambda)$. 
\end{assumption}
For a reversible chain, $1-\lambda$ is the absolute spectral gap. The multiplier $\tau_\lambda$ bounds dependence over this contraction class and equals one in the i.i.d.\ case. The limiting covariance of the estimator depends on the Hessian of the population loss at its minimizer and on the long-run covariance of the scores.

Local estimation and inference take place in a neighborhood of the population minimizer where the Riemannian Hessian of the population loss is positive definite. For $x\in\M$, let $T_x\M$ denote the tangent space at $x$, with inner product $\left\langle\cdot,\cdot\right\rangle_x$ and norm $\norm{\cdot}_x$ induced by $g$ (omitting the base-point subscript when clear from context). Let $D_x\subset T_x\M$ consist of the initial velocities $v$ for which the geodesic starting at $x$ with velocity $v$ exists in $\M$ throughout $\left[0,1\right]$, including its endpoints. The exponential map $\Exp_x:D_x\to\M$ sends $v$ to the endpoint of that geodesic. For a smooth function on $\M$, $\grad$ and $\Hess$ denote its Riemannian gradient and Hessian, with $\norm{\cdot}_\op$ denoting the induced tangent-space operator norm. For points $x$ and $y$ joined by a unique minimizing geodesic, $\cP_{x\to y}:T_x\M\to T_y\M$ denotes parallel transport along that geodesic.

Let $x^*$ be a minimizer of $f$. For $s>0$, write $B_s(0)=\{v\in T_{x^*}\M:\norm{v}<s\}$ for the tangent ball and $U_s=\{x\in\M:d(x,x^*)<s\}$ for the geodesic ball. For $s$ below the local convexity radius at $x^*$, parallel transport between points of $\overline U_s$ follows their unique minimizing geodesic in $\overline U_s$. 

\begin{assumption}
\label{ass:local}
The minimizer $x^*$ is unique and lies in $\operatorname{int}\M$. Choose $r>0$ below the local convexity radius at $x^*$ such that $\Exp_{x^*}$ sends an open neighborhood of $\overline{B_r(0)}$ diffeomorphically onto an interior normal neighborhood containing $\overline U_r$, with $\Exp_{x^*}(B_r(0))=U_r$. For each data state $z\in\Xi$, the map $x\mapsto\ell(x,z)$ is twice continuously differentiable on the normal neighborhood. There are constants $K_1,K_2,K_3<\infty$ such that, for every $z\in\Xi$,
\begin{align}
  \norm{\grad\ell(x^*,z)}&\le K_1,
  \label{eq:score_bound_intrinsic}\\
  \norm{\Hess\ell(x,z)}_\op&\le K_2,
  \quad x\in\overline U_r,
  \label{eq:hessian_bound_intrinsic}\\
  \norm{
    \cP_{y\to x}\Hess\ell(y,z)\cP_{x\to y}
    -\Hess\ell(x,z)
  }_\op
  &\le K_3d(x,y),
  \quad x,y\in\overline U_r.
  \label{eq:hessian_transport_lipschitz}
\end{align}
The population Hessian $\mathsf{H}:=\Hess f(x^*):T_{x^*}\M\to T_{x^*}\M$ is positive definite, with minimum eigenvalue $\lambda_*:=\lambda_{\min}(\mathsf{H})>0$, and
\begin{equation}
  K_3r/2\le\lambda_*/4.
  \label{eq:local_radius_condition}
\end{equation}
\end{assumption}
Because $\lambda_*>0$ and the same Hessian Lipschitz bound $K_3$ holds on smaller balls, we can decrease $r$ to satisfy \eqref{eq:local_radius_condition}. For the chosen radius, define the identification gap $\gamma_0:=\inf_{x\notin U_{r/2}}\{f(x)-f(x^*)\}$. It is positive by compactness of $\M\setminus U_{r/2}$, continuity of $f$, and uniqueness of $x^*$. 

Here $K_1\le G$ captures the local score scale at $x^*$. The transported Hessian condition \eqref{eq:hessian_transport_lipschitz} is coordinate-free. For $C^3$ losses,  it follows from a bound on the covariant derivative of $\Hess\ell(\cdot,z)$ that is uniform in $z$ throughout the neighborhood containing the transport geodesics.  Assumption~\ref{ass:spectral} employs the $L^2_0(\pi)$ operator-norm contraction underlying the exponential concentration inequalities in \citet{fanjiangsun2021} for bounded observables of a stationary Markov chain. The uniform-in-$z$ bounds in Assumptions~\ref{ass:global} and~\ref{ass:local} allow direct application of these inequalities to the loss and its local derivatives. The local smoothness and positive-definite Hessian in Assumption~\ref{ass:local} parallel the differentiability and nonsingularity conditions in the intrinsic-mean limit theory of \citet{bhattacharya2005}. The uniform derivative bounds provide quantitative control of the error at finite sample sizes. 

Two canonical geometric estimation problems illustrate how these abstract constants encode cut-locus separation or an eigengap. For a Fr\'echet-type objective on the unit sphere $\mathbb S^m$, take $\ell(x,z)=\tfrac12 d(x,z)^2$. Since $\operatorname{diam}(\mathbb S^m)=\pi$, the loss satisfies $0\le \ell(x,z)\le \frac{\pi^2}{2}$, and $
|\ell(x,z)-\ell(y,z)|\le \pi d(x,y).$ Thus Assumption~\ref{ass:global} holds globally on $\mathbb S^m\times\mathbb S^m$ with $B=\pi^2/2$ and $G=\pi$, despite the nonsmoothness of $x\mapsto\ell(x,z)$ at the antipodal cut locus $x=-z$. Uniform separation of the observation support $\Xi$ from the cut loci of points in the normal neighborhood containing $\overline U_r$ ensures smoothness and uniform derivative bounds on $\overline U_r\times\Xi$. A sufficiently small geodesic ball containing the observation support ensures uniqueness of the population Fr\'echet mean and positive definiteness of the population Hessian in a neighborhood of that mean \citep{afsari2011}. %


For principal-subspace estimation, let $\mathrm{Gr}(q,p)$ denote the Grassmann manifold of $q$-dimensional subspaces of $\R^p$, where $1\le q<p$. Represent $x\in\mathrm{Gr}(q,p)$ by an orthonormal frame $\mathbf X\in\R^{p\times q}$ and use
\begin{equation}
 \ell(x,z)=-\tr(\mathbf X^\top zz^\top\mathbf X),\qquad z\in\R^p.
 \label{eq:grassmann_loss}
\end{equation}
The loss depends on $\mathbf X$ only through $\mathbf X\mathbf X^\top$ and is therefore well defined on $\mathrm{Gr}(q,p)$. We use the canonical horizontal metric $\langle\Delta_1,\Delta_2\rangle=\tr(\Delta_1^\top\Delta_2)$; its geodesic distance is the Euclidean norm of the $q$ principal angles. When $\Xi$ is bounded, the loss and its derivatives are uniformly bounded on the compact Grassmann manifold. A strict gap between the $q$th and $(q+1)$st eigenvalues of $\E_\pi[\xi\xi^\top]$ identifies the leading subspace and makes the population Hessian positive definite in its tangent space.


\subsection{Computed estimators}\label{sec:computed-estimators}

Numerical optimization algorithms return measurable approximate solutions of $L_N$. We use two standard accuracy criteria. 
\begin{definition}
\label{def:approx}
Let $\wt x_N\in\M$ be a measurable computed estimator, and let $\eps_N^{\rm obj},\eps_N^{\rm grad}\ge0$ be accuracy tolerances. An $\eps_N^{\rm obj}$-approximate SAA solution satisfies
\begin{equation}
  L_N(\wt x_N)\le \inf_{x\in\M}L_N(x)+\eps_N^{\mathrm{obj}}.
  \label{eq:objective_certificate}
\end{equation}
An $\eps_N^{\rm grad}$-stationary solution is a point $\wt x_N$ at which $L_N$ is differentiable and satisfies
\begin{equation}
  \norm{\grad L_N(\wt x_N)}\le\eps_N^{\mathrm{grad}}.
  \label{eq:gradient_certificate}
\end{equation}
\end{definition}

In the finite-sample results, the objective-gap and gradient tolerances are deterministic nonnegative numbers, though the bounds remain valid if random tolerances are almost surely dominated by them. In the asymptotic corollaries, random tolerances are accommodated via $o_{\Pbb}$ rates together with high-probability localization.


While an objective-gap certificate measures suboptimality against the global empirical minimum, a gradient residual is computable purely from local derivatives once an iterate lies within $U_{r/2}$. We therefore state localization conditions explicitly in the corresponding guarantees.
A local correction uses the empirical gradient and Hessian at the same point. Specifically, given a pilot $\bar x_N\in\M$ and a linear-solve tolerance $\eta_N\ge0$, we seek a tangent direction $v_N^{\mathrm{os}}\in T_{\bar x_N}\M$ satisfying %
\begin{equation}
  \norm{
    \Hess L_N\left(\bar x_N\right)v_N^{\rm os}
    +\grad L_N\left(\bar x_N\right)
  }
  \le\eta_N,
  \label{eq:one_step_direction_final}
\end{equation}
which can be computed via matrix-free solvers such as truncated conjugate gradients on $T_{\bar x_N}\M$ using only empirical Hessian-vector products. The exact Riemannian Newton step corresponds to $\eta_N=0$. Let $\mathcal A_N^{\mathrm{os}}$ denote the event that $x\mapsto\ell(x,\xi_i)$ is twice differentiable in a neighborhood of $\bar x_N$ for all $i=1,\ldots,N$ and the linear solver returns a measurable direction $v_N^{\mathrm{os}}\in D_{\bar x_N}$ satisfying \eqref{eq:one_step_direction_final}. 
Define the one-step estimator by $x_N^{\mathrm{os}}=\Exp_{\bar x_N}(v_N^{\mathrm{os}})$ on $\mathcal A_N^{\mathrm{os}}$ and $x_N^{\mathrm{os}}=x_{\mathrm{ref}}$ on $(\mathcal A_N^{\mathrm{os}})^c$ for a fixed $x_{\mathrm{ref}}\in\M$. Section~\ref{sec:one-step} gives sufficient accuracy conditions for this update.


\subsection{Covariance estimation and confidence regions}
\label{sec:inference}
For the Wald confidence region, let $x_N^\dagger$ denote the selected estimator output, such as a direct approximation, the one-step estimate, or exact SAA. The population score $\zeta(z)=\grad\ell(x^*,z)$ takes values in $V=T_{x^*}\M$ and has mean zero. For $h=0,1,\ldots$, write $\mathsf C_h=\E_\pi[\zeta(\xi_1)\ot\zeta(\xi_{1+h})]$, where $(a\ot b)v=a\inner{b}{v}$. Letting $\mathsf C_h^*$ denote the adjoint of $\mathsf C_h$ with respect to $\langle\cdot,\cdot\rangle_{x^*}$, the long-run score covariance is $\mathsf S=\mathsf C_0+\sum_{h=1}^{\infty}(\mathsf C_h+\mathsf C_h^*)$ and the target parameter covariance is $\mathsf V_*=\mathsf H^{-1}\mathsf S\mathsf H^{-1}$. The series converges absolutely under spectral contraction, and $\mathsf V_*$ is positive definite whenever $\mathsf S\succ0$, as imposed in Section~\ref{sec:inference-validity}. Its lag terms describe the serial uncertainty within the given model, while $\tau_\lambda$ provides a class-wide bound.

Choose an integer batch length $1\le b_N\le N$ and set $a_N=\left\lfloor N/b_N\right\rfloor$.
Let $\mathcal D_N$ denote the event that $a_N\ge2$ and the sample gradients and Hessians can be evaluated at $x_N^\dagger$
for all $i=1,\ldots,N$.
On $\mathcal D_N$, define $Y_i^{(N)}=\grad\ell(x_N^\dagger,\xi_i)\in T_{x_N^\dagger}\M$
and form the complete-batch means $\bar Y_j^{(N)}=b_N^{-1}\sum_{i=(j-1)b_N+1}^{jb_N}Y_i^{(N)}$ for $j=1,\ldots,a_N$.
With
$\bar Y^{(N)}=a_N^{-1}\sum_{j=1}^{a_N}\bar Y_j^{(N)}$, the batch-means covariance is
\[
\wh{\mathsf S}_N^\dagger
=
\frac{b_N}{a_N-1}
\sum_{j=1}^{a_N}
\left(\bar Y_j^{\left(N\right)}-\bar Y^{\left(N\right)}\right)
\ot
\left(\bar Y_j^{\left(N\right)}-\bar Y^{\left(N\right)}\right).
\]
The first $a_Nb_N$ observations enter this covariance estimate, while the empirical Hessian $\wh{\mathsf H}_N^\dagger=\Hess L_N(x_N^\dagger)$ uses all $N$ observations; on $\mathcal D_N^c$, set $\wh{\mathsf S}_N^\dagger$ and $\wh{\mathsf H}_N^\dagger$ equal to the identity operator on $T_{x_N^\dagger}\M$.
Let $\mathcal A_N$ be the event that $\mathcal D_N$ holds, $\wh{\mathsf H}_N^\dagger$ is invertible, and $\wh{\mathsf S}_N^\dagger\succ0$. On $\mathcal A_N$, form the sandwich operator $\wh{\mathsf V}_N^\dagger=(\wh{\mathsf H}_N^\dagger)^{-1}\wh{\mathsf S}_N^\dagger(\wh{\mathsf H}_N^\dagger)^{-1}$.
On $\mathcal A_N^c$, set $\wh{\mathsf V}_N^\dagger$ equal to
the identity operator on $T_{x_N^\dagger}\M$.

For $0<\alpha<1$, let $q_{m,1-\alpha}$ be the $\left(1-\alpha\right)$ quantile of the chi-squared distribution with $m$ degrees of freedom.
On $\mathcal A_N$, define the estimated tangent ellipsoid
\[
\mathcal E_{N,1-\alpha}^\dagger
=
\left\{
v\in T_{x_N^\dagger}\M:
N\left\langle
v,\left(\wh{\mathsf V}_N^\dagger\right)^{-1}v
\right\rangle_{x_N^\dagger}
\le q_{m,1-\alpha}
\right\}.
\]
and set $\mathcal E_{N,1-\alpha}^\dagger=\varnothing$ on $\mathcal A_N^c$. The reported confidence region is
\begin{equation}
\mathcal C_{N,1-\alpha}^\dagger
=
\left\{
\Exp_{x_N^\dagger}\left(v\right):
v\in\mathcal E_{N,1-\alpha}^\dagger\cap D_{x_N^\dagger}
\right\}.
\label{eq:wald_region_definition}
\end{equation}
In numerical implementation, fixing any $g_{x_N^\dagger}$-orthonormal basis of $T_{x_N^\dagger}\M$ represents $\wh{\mathsf S}_N^\dagger$, $\wh{\mathsf H}_N^\dagger$, and $\wh{\mathsf V}_N^\dagger$ as $m\times m$ symmetric matrices, avoiding ambient-space rank deficiency on embedded manifolds such as $\mathrm{Gr}(q,p)$. The construction uses the computed estimator, empirical derivatives, and batch-means covariance from the same trajectory. 
Section~\ref{sec:inference-validity} establishes the asymptotic coverage of this region.

\section{Theoretical Results}
\label{sec:theory}
\subsection{Estimation error, risk, and limiting distribution}
\label{sec:local}

Given a single trajectory of length $N$, how accurately should we solve the empirical problem? Even its exact minimizer varies from one trajectory to the next, and serial dependence affects the size of that variation. A finite-sample bound on the distance to $x^*$ lets us compare the error allowed by the solver's tolerance with the sampling error at the available trajectory length.

Write $D_\M=\diam(\M)$ and let $\cN(\M,d,t)$ be the geodesic covering number. Define $\cJ_\M=\int_0^{D_\M}\sqrt{\log\{2\cN(\M,d,t)\}}\,\mathrm{d}t$. For $u\in(0,1)$, set
\begin{equation}
\begin{aligned}
 \mathfrak D_N(u)&=C_{\rm ch}\sqrt{\frac{\tau_\lambda}{N}}
 \left[G\cJ_\M+(GD_\M+B)\sqrt{\log(4/u)}\right],\\
 \bigl(s_N(u),h_N(u)\bigr)&=5(K_1,K_2)\sqrt{\frac{\tau_\lambda}{N}
                 \left(m+\log(2/u)\right)},
\end{aligned}
\label{eq:concentration_radii}
\end{equation}
where $C_{\rm ch}$ is universal. For $\delta\in(0,1)$, put $\delta_0=\delta/3$ and abbreviate $s_N=s_N(\delta_0)$ and $h_N=h_N(\delta_0)$. These quantities control the uniform objective, score, and Hessian fluctuations, respectively.

\begin{theorem}
\label{thm:local_fast_rate}
Under Assumptions~\ref{ass:global}--\ref{ass:local}, let $\wh x_N$ be an exact global SAA minimizer. If $h_N\le\lambda_*/4$ and $2\mathfrak D_N\left(\delta_0\right)<\gamma_0$, the following bounds hold on a common event of probability at least $1-\delta$. Any $\wt x_N$ satisfying \eqref{eq:objective_certificate} with $2\mathfrak D_N\left(\delta_0\right)+\eps_N^{\rm obj}<\gamma_0$ lies in $U_{r/2}$ and satisfies
\[
 d\left(\wt x_N,x^*\right)\le\frac{2s_N}{\lambda_*}+2\sqrt{\frac{\eps_N^{\rm obj}}{\lambda_*}}.
\]
In particular, $\wh x_N\in U_{r/2}$ and its bound uses $\eps_N^{\rm obj}=0$. Any $\wt x_N^{\rm grad}\in U_{r/2}$ satisfying \eqref{eq:gradient_certificate} obeys $d(\wt x_N^{\rm grad},x^*)\le 2(s_N+\eps_N^{\rm grad})/\lambda_*$.
\end{theorem}

Write $g_N=N^{-1}\sum_{i=1}^N\zeta\left(\xi_i\right)$ for the average score at $x^*$ and $\mathsf S_N=\Cov(\sqrt N\,g_N)$ for its scaled covariance. The finite-trajectory covariance $\mathsf S_N$ converges to $\mathsf S$. On $U_r$, $\Log_{x^*}$ denotes the inverse of $\Exp_{x^*}|_{B_r\left(0\right)}$, extended by zero outside $U_r$.

\begin{theorem}
\label{thm:asymptotic_normality}
Under Assumptions~\ref{ass:global}--\ref{ass:local}, with fixed dimension, contraction, and regularity constants, a measurable exact global SAA minimizer satisfies $
 \E d\left(\wh x_N,x^*\right)^2
 =N^{-1}\tr\left(\mathsf H^{-1}\mathsf S_N\mathsf H^{-1}\right)+o\left(N^{-1}\right),
 \E\left\{f\left(\wh x_N\right)-f\left(x^*\right)\right\}
 =(2N)^{-1}\tr\left(\mathsf H^{-1}\mathsf S_N\right)+o\left(N^{-1}\right). $ 
The tangent displacement satisfies
\begin{equation}
  \sqrt N\,\Log_{x^*}\left(\wh x_N\right)
  =-\mathsf H^{-1}\sqrt N\,g_N+o_{\Pbb}\left(1\right)
  \xrightarrow{d}\mathcal N_V\left(0,\mathsf V_*\right).
  \label{eq:asymptotic_linearity_final}
\end{equation}

If $Z\sim\mathcal N_V\left(0,\mathsf{S}\right)$, then
\begin{equation}
  N\left\{f\left(\wh x_N\right)-f\left(x^*\right)\right\}
  \xrightarrow{d}
  \frac12\inner{Z}{\mathsf{H}^{-1}Z}.
  \label{eq:second_order_risk_limit_final}
\end{equation}
\end{theorem}

Squared geodesic error sums the squared components of the tangent displacement in an orthonormal eigenbasis of the population Hessian $\mathsf H$. To second order, excess loss weights each squared component by half the corresponding eigenvalue. Both distributional limits allow a singular long-run score covariance.

\subsection{Minimax rates on compact manifolds}
\label{sec:lower_bound}

The preceding risk expansions describe exact SAA for a given loss and chain. To determine whether another estimator can improve the squared-error rate uniformly, we compare exact SAA with all trajectory-based estimators on a model class fixed across sample sizes. 
For $\theta\in[-1/4,1/4]^m$, let $\pi_\theta$ be the product law on $\{-1,+1\}^m$ with mean $\theta$. Fix $0\le\lambda<1$ and consider stationary chains with invariant law $\pi_\theta$ and centered $L^2(\pi_\theta)$ contraction at most $\lambda$. Write $\mathfrak R_{N,m}^{\M}(r,\lambda)$ for the minimax squared geodesic risk over this class; Appendix~\ref{sec:general_manifold_minimax} gives its formal definition and the loss construction.

\begin{theorem}[Minimax rate on compact manifolds]
\label{thm:general_manifold_minimax}
On any compact $m$-dimensional Riemannian manifold ($m\ge1$) with possibly empty smooth boundary, choose a sufficiently small normal chart of radius $2r$ with closure in the interior. There exists a common bounded smooth loss, fixed across sample sizes, with unique population and exact SAA minimizers $x_\theta^*$ and $\wh x_N^{\rm ch}$ in the chart. For every $N\ge1$,
\begin{equation}
 \mathfrak R_{N,m}^{\M}\left(r,\lambda\right)
 \asymp \sup_{\theta,\mathsf P}\E_{\theta,\mathsf P}
 d\left(\wh x_N^{\rm ch},x_\theta^*\right)^2
 \asymp r^2\min\left\{1,\frac{m\tau_\lambda}{N}\right\}.
 \label{eq:general_minimax_rate}
\end{equation}
The comparison constants depend only on the chart metric bounds.
\end{theorem}

For $N/\tau_\lambda\ge m$, the optimal squared-error order decreases as $r^2m\tau_\lambda/N$. For shorter effective trajectories, the minimax risk is of order $r^2$, the squared spatial scale of the parameter class.

\subsection{Numerical accuracy and one-step equivalence}
\label{sec:one-step}

Exact SAA provides a statistical benchmark, but in practice the estimator is the point at which the solver stops. Two estimators can both have error of order $N^{-1/2}$ and still have different limits after scaling by $\sqrt N$. To use the sampling law of exact SAA for the computed estimate, it suffices to make their difference negligible on the $N^{-1/2}$ scale.

One option is to return the iterate directly. Its accuracy can be expressed through the objective gap relative to the global empirical minimum or through the gradient residual at the iterate. The gradient residual can be evaluated at the iterate itself.

A second option is to use the iterate as a pilot. Using the same trajectory, we evaluate the empirical gradient and Hessian at the pilot, solve the Newton equation in its tangent space, and map the step back to the manifold with the exponential map. Near the empirical minimizer the remaining error is bounded by a term quadratic in the pilot's distance to that minimizer and a term linear in the solve residual. This permits a less accurate pilot. For the direct and one-step estimators defined in Section~\ref{sec:computed-estimators}, the sufficient tolerances are summarized below. The relaxed pilot requirements follow from the quadratic contraction in Theorem~\ref{thm:one_step_quadratic}.


The two columns below give alternative sufficient tolerances for \eqref{eq:objective_certificate} and \eqref{eq:gradient_certificate}. For the gradient criterion, the direct estimate or pilot must lie in $U_{r/2}$ with probability tending to one, and the gradient bound must hold there.
\begin{table}[htbp]
  \centering
  \caption{Asymptotic tolerance requirements.}
  \label{eq:asymptotic_gradient_localization}
  \begin{tabular}{lcc}
    \hline
    & Objective-gap tolerance & Gradient tolerance \\
    \hline
    Direct estimate $\wt{x}_N$
      & $o_{\Pbb}\left(N^{-1}\right)$
      & $o_{\Pbb}\left(N^{-1/2}\right)$ \\
    Newton pilot $\bar{x}_N$
      & $o_{\Pbb}\left(N^{-1/2}\right)$
      & $o_{\Pbb}\left(N^{-1/4}\right)$ \\
    \hline
  \end{tabular}
\end{table}
For either pilot criterion, require the linear solve in \eqref{eq:one_step_direction_final} to have tolerance
\begin{equation}
 \eta_N=o_{\Pbb}\left(N^{-1/2}\right).
 \label{eq:inexact_solve_threshold}
\end{equation}

\begin{corollary}
\label{cor:one_step_equivalence}
Under Assumptions~\ref{ass:global}--\ref{ass:local}, fix the dimension, contraction, and regularity constants. Let $x_N^\dagger$ be the direct or one-step estimate satisfying its respective accuracy conditions above. The Newton update, when used, is feasible and lies in $U_r$ with probability tending to one. For either estimate,
\begin{equation}
 \sqrt N\left\{\Log_{x^*}\left(x_N^\dagger\right)-\Log_{x^*}\left(\wh x_N\right)\right\}\xrightarrow{p}0.
 \label{eq:one_step_equivalence}
\end{equation}
Hence both estimates have the linear representation and the parameter and excess-risk distributional limits of Theorem~\ref{thm:asymptotic_normality}.
\end{corollary}

For example, consider a localized iterate computed to a gradient tolerance of $N^{-1/3}$. This tolerance is too coarse for the direct-estimate criterion but meets the pilot criterion. With a linear solve satisfying \eqref{eq:inexact_solve_threshold}, one Newton step from this iterate gives an estimate with the same first-order distribution as exact SAA.

\subsection{Validity of confidence regions}
\label{sec:inference-validity}
To turn the limiting distribution into a confidence region, we estimate $\mathsf V_*$ using the empirical Hessian and a batch-means estimate of the long-run score covariance. We consider any output satisfying $\Pbb\{x_N^\dagger\in U_r\}\to1$ and, with the local logarithm set to zero on its complement,
\begin{equation}
 \sqrt N\,\Log_{x^*}(x_N^\dagger)
 =-\mathsf H^{-1}\frac1{\sqrt N}\sum_{i=1}^N\zeta\left(\xi_i\right)
 +o_{\Pbb}\left(1\right).
 \label{eq:generic_asymptotic_linearity}
\end{equation}
We estimate the population loss Hessian by the empirical Hessian and long-run score covariance by batch means, both from the trajectory used to fit the estimator. For the batch estimate we impose the following conditions on the chain and batch length.

\begin{assumption}
\label{ass:batch_means}
The stationary chain $\left(\xi_i\right)_{i\ge1}$ is $\pi$-Harris recurrent and geometrically ergodic in total variation. The long-run score covariance $\mathsf S$ is positive definite. For a fixed $\nu\in\left(1/2,1\right)$, use the batch length $b_N=\left\lfloor N^\nu\right\rfloor$.
\end{assumption}

The empirical Hessian and batch-means covariance are evaluated at the computed estimate, so they act on $T_{x_N^\dagger}\M$. To compare them with their population counterparts, transport them to $T_{x^*}\M$ when $x_N^\dagger\in U_{r/2}$, and set the differences to zero otherwise.

\begin{assumption}
\label{ass:batch_means}
The stationary chain $\left(\xi_i\right)_{i\ge1}$ is $\pi$-Harris recurrent and geometrically ergodic in total variation. The long-run score covariance $\mathsf S$ is positive definite. For a fixed $\nu\in\left(1/2,1\right)$, use the batch length $b_N=\left\lfloor N^\nu\right\rfloor$.
\end{assumption}

The empirical Hessian and batch-means covariance are evaluated at the computed estimate, so they act on $T_{x_N^\dagger}\M$. To compare them with their population counterparts, transport them to $T_{x^*}\M$ when $x_N^\dagger\in U_{r/2}$, and set the differences to zero otherwise.

\begin{theorem}[Transported covariance consistency]
\label{thm:plugin_covariance}
Under Assumptions~\ref{ass:global}--\ref{ass:local} and~\ref{ass:batch_means}, suppose $x_N^\dagger$ satisfies \eqref{eq:generic_asymptotic_linearity} and $\Pbb\{x_N^\dagger\in U_r\}\to1$, with fixed dimension, contraction, and regularity constants. After transport, the empirical Hessian, batch-means score covariance, and sandwich covariance converge in probability in operator norm to $\mathsf H$, $\mathsf S$, and $\mathsf V_*$, respectively.
\end{theorem}

\begin{corollary}
\label{cor:wald_region}
Under Assumptions~\ref{ass:global}--\ref{ass:local} and~\ref{ass:batch_means}, fix the dimension, contraction, and regularity constants. Suppose $x_N^\dagger$ satisfies \eqref{eq:generic_asymptotic_linearity} and $\Pbb\{x_N^\dagger\in U_r\}\to1$. Then for each $0<\alpha<1$ the region in \eqref{eq:wald_region_definition} satisfies $\Pbb\{x^*\in\mathcal C_{N,1-\alpha}^\dagger\}\longrightarrow1-\alpha$.
\end{corollary}

\section{Simulations}
\label{sec:numerical-additional}


\subsection{Simulation setup}
We simulate stationary Markov chains $(\xi_i)_{i=1}^N$ on $\{-1,+1\}^m$ with invariant product law $\pi_\theta$ under two transition mechanisms sharing the same $L^2_0(\pi_\theta)$ contraction $\lambda$: a whole-vector refresh kernel (retaining $\xi_{i-1}$ with probability $\lambda$ and redrawing from $\pi_\theta$ with probability $1-\lambda$) and a coordinatewise refresh kernel. We pair these chains with smooth bounded losses on the flat torus $\mathbb T^m$ and the unit sphere $\mathbb S^m$. Exact loss specifications and tuning protocols are detailed in Appendix~\ref{sec:simulation-design}.
To evaluate estimation error and verify Theorems~\ref{thm:local_fast_rate}-\ref{thm:general_manifold_minimax}, we vary trajectory length $N$, dimension $m$, and persistence $\lambda$. To isolate numerical error around the thresholds of Corollary~\ref{cor:one_step_equivalence}, we synthetically perturb the empirical minimizer $\wh x_N$ at controlled objective-gap and gradient-residual rates and apply one intrinsic Newton step \eqref{eq:one_step_direction_final} with controlled linear-solve residuals $\eta_N$.
For inference, we select the batch length $b_N$ on independent tuning trajectories and evaluate the resulting plug-in batch-means Wald regions \eqref{eq:wald_region_definition} across $1000$ independent evaluation trajectories per setting.


\subsection{Results}

\begin{figure}[htbp]
\centering
\includegraphics[width=\textwidth,trim=29bp 26bp 10bp 18bp,clip]{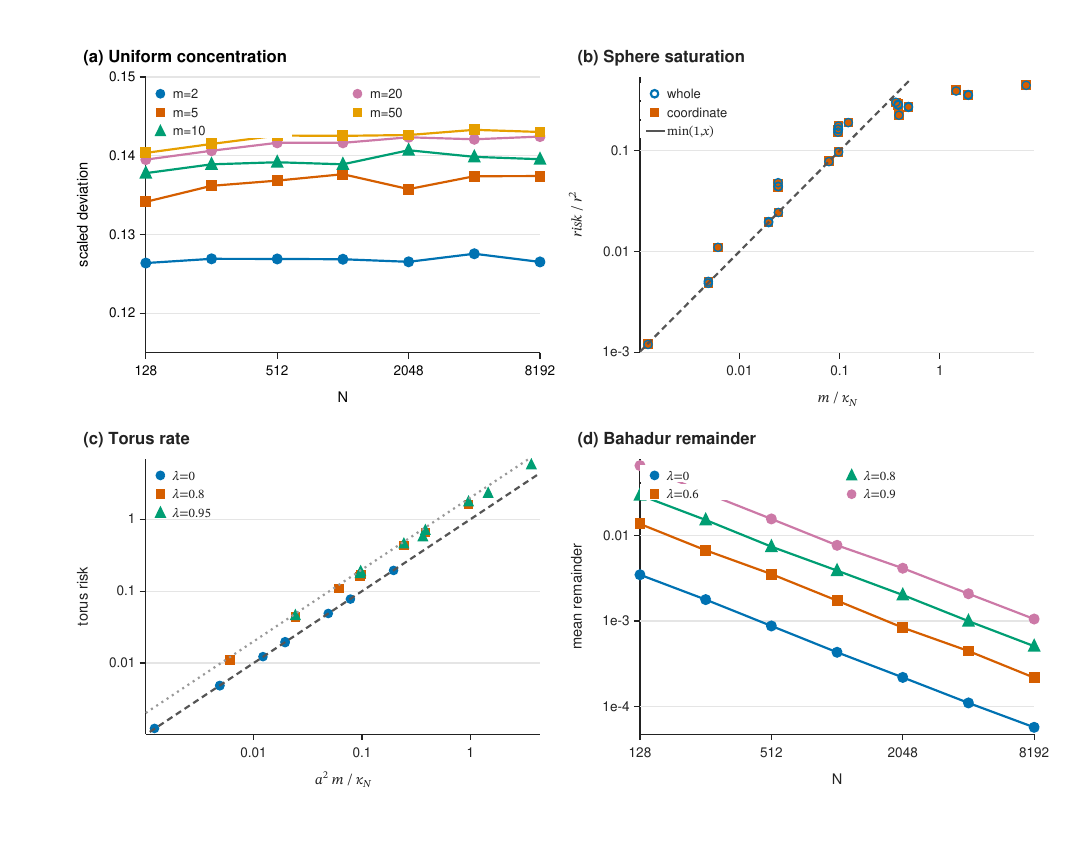}
\caption{Dependence and sample-size scaling across manifold geometries. (a) Normalized uniform objective deviations on $\mathbb S^m$. (b) Normal-chart sphere risk under whole-vector and coordinatewise refresh kernels compared with $\min\{1,m/\kappa_N\}$, where $\kappa_N=1+(N-1)(1-\lambda)$ {proxies the effective sample size}. (c) Torus squared geodesic risk against {$m/\kappa_N$ (unit loss scale)}. (d) Mean {unscaled} intrinsic Bahadur remainder {$\norm{\Log_{x^*}(\wh x_N)+\mathsf H^{-1}g_N}$} for $m=5$. Reference lines indicate theoretical rates.}

\label{fig:rate-results}
\end{figure}

{Figure~\ref{fig:rate-results} corroborates the finite-sample and asymptotic estimation theory of Section~\ref{sec:theory}.} In panel~(a), the normalized uniform objective deviations remain stable across $N$, $m$, and $\lambda$, ranging tightly from $0.123$ to $0.144$ {and confirming the $\sqrt{\tau_\lambda/N}$ empirical-process scaling in $\mathfrak D_N$}. Consistent with the minimax transition in Theorem~\ref{thm:general_manifold_minimax}, the sphere risk levels off for short effective trajectories and decays at the $m/\kappa_N$ rate once $\kappa_N\gtrsim m$ (panel~b). Moreover, the whole-vector and coordinatewise refresh kernels yield nearly identical risks (mean ratio $0.998$), showing that the spectral parameter $\lambda$ captures the dominant serial effect. On the torus, mean squared geodesic error averages $0.963$ times the finite-trajectory variance bound used in the torus experiment (panel~c). In panel~(d), the unscaled linearization remainder $\norm{\Log_{x^*}(\wh x_N)+\mathsf H^{-1}g_N}$ has a median log-log slope of $-0.989$, matching the $O_{\Pbb}(N^{-1})$ Bahadur remainder underlying \eqref{eq:asymptotic_linearity_final}.

\begin{figure}[htbp]
\centering
\includegraphics[width=\textwidth,trim=30bp 31bp 8bp 19bp,clip]{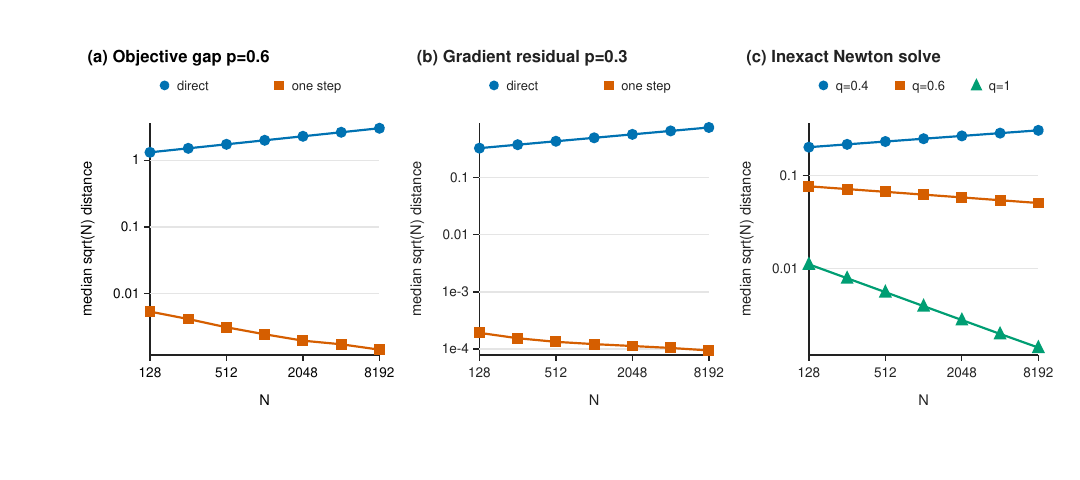}
\caption{Numerical accuracy relative to exact SAA, measured by
$\sqrt N\,d(\cdot,\wh x_N)$. Panels (a) and (b) compare controlled
objective-gap and gradient-residual pilots with one exact Newton correction.
Panel (c) varies the linear-system residual exponent $q$ with $\eta_N\asymp N^{-q}$.}
\label{fig:supp-one-step}
\end{figure}
Figure~\ref{fig:supp-one-step} {validates the sharp tolerance thresholds of Corollary~\ref{cor:one_step_equivalence}} by comparing each pilot $\bar x_N$ and its one-step update $x_N^{\mathrm{os}}$ against $\wh x_N$. At pilot accuracies between the direct and one-step regimes {in Table  \ref{eq:asymptotic_gradient_localization} (between $N^{-1}$ and $N^{-1/2}$ for $\eps_N^{\mathrm{obj}}$, and between $N^{-1/2}$ and $N^{-1/4}$ for $\eps_N^{\mathrm{grad}}$)}, the scaled discrepancy $\sqrt N\,d(\bar x_N,\wh x_N)$ diverges for the uncorrected pilot whereas $\sqrt N\,d(x_N^{\mathrm{os}},\wh x_N)$ vanishes after a single Newton step (panels~a and~b). In panel~(c), varying $\eta_N\asymp N^{-q}$ confirms the $q=1/2$ boundary in \eqref{eq:inexact_solve_threshold}: the post-step root-$N$ discrepancy grows for $q=0.4$ and decays to zero for $q=0.6$ and $1.0$.

\begin{table}[htbp]
\centering
\caption{Coverage of nominal 95\% $F$-type feasible Wald regions on independent
evaluation trajectories. }
\small
\label{tab:independent-coverage}
\begin{tabular}{cccccc}
\toprule
Manifold & $\lambda$ & $N=1000$ & $N=4000$ & $N=16000$ & $N=64000$\\
\midrule
$\mathbb T^2$ & 0 & 0.946 & 0.950 & 0.949 & 0.939\\
$\mathbb T^2$ & 0.8 & 0.921 & 0.945 & 0.955 & 0.945\\
$\mathbb T^2$ & 0.95 & 0.791 & 0.912 & 0.927 & 0.946\\
$\mathbb S^2$ & 0 & 0.955 & 0.955 & 0.955 & 0.933\\
$\mathbb S^2$ & 0.8 & 0.921 & 0.929 & 0.938 & 0.954\\
$\mathbb S^2$ & 0.95 & 0.827 & 0.917 & 0.917 & 0.959\\
\bottomrule
\end{tabular}
\end{table}
Table~\ref{tab:independent-coverage} reports empirical coverage {of the Corollary~\ref{cor:wald_region} confidence regions for $m=2$ ($\mathbb T^2$ and $\mathbb S^2$)} across $1000$ independent evaluation trajectories per setting, calibrated with the Hotelling cutoff $q_{2,a_N,0.95}=2(a_N-1)F_{2,a_N-2;\,0.95}/(a_N-2)$. Under strong persistence $\lambda=0.95$, where $\tau_\lambda=39$ leaves only $N/\tau_\lambda\approx25.6$ effective observations at $N=1000$, finite-sample coverage starts at $0.791$ on $\mathbb T^2$ and $0.827$ on $\mathbb S^2$, then improves steadily as the effective sample size grows. At $N=64000$, coverage ranges from $0.933$ to $0.959$ across all six configurations.

\section{Compressor operating subspaces in MetroPT-3}
\label{sec:metropt}

\subsection{Data and chronological design}
\label{sec:metropt-details}

To illustrate the methodology on autocorrelated industrial telemetry, we study onboard air-compressor sensor streams from an operational metro train in the UCI MetroPT-3 dataset \citep{davari2021metropt}, recorded at $10$-second intervals ($0.1\,\mathrm{Hz}$) across seven analog pressure, temperature, and motor-current channels. Because charging and blowdown cycles couple these sensors into a low-dimensional operating mode, tracking the dominant subspace provides a geometric summary of compressor regime shifts. Using February~2020 data to fix standardization parameters and solver initialization, we analyze seven consecutive pre-fault windows from March~2 through April~17, 2020, comprising $340{,}451$ observations in total.  We use the analog channels TP2, TP3, H1, DV\_pressure, Reservoirs, Oil\_temperature, and Motor\_current.



\subsection{Grassmann criterion and computation}
\label{sec:metropt-computation}

For each analog channel, we subtract its February mean, divide by its February standard deviation (with the variance divisor equal to the February sample size), and clip to $[-5,5]$. Clipping limits the influence of telemetry spikes and affects 1.75\% of evaluation entries. The resulting support $\Xi=[-5,5]^7$ gives uniform bounds on the smooth Grassmann loss and its derivatives, as required by the boundedness conditions in Assumptions~\ref{ass:global} and~\ref{ass:local}. The target is the dominant subspace of this bounded, transformed second moment.

Within each window, let $N$ be the number of transformed readings $z_1,\ldots,z_N$. We specialize \eqref{eq:grassmann_loss} to $\mathrm{Gr}(2,7)$, of intrinsic dimension $m=2(7-2)=10$, using an orthonormal frame $\mathbf X\in\R^{7\times2}$ to represent the subspace $[\mathbf X]$. The canonical horizontal inner product is $\langle\Delta_1,\Delta_2\rangle=\tr(\Delta_1^\top\Delta_2)$, and geodesic distance is the Euclidean norm of the two principal angles. In this application,
\[
 \ell(\mathbf X,z)
 =-\|\mathbf X^\top z\|^2
 =-\tr(\mathbf X^\top zz^\top\mathbf X),
 \qquad
 \mathbf C_N=\frac1N\sum_{i=1}^N z_i z_i^\top.
\]
The second moment is uncentered within each window, since preprocessing is fixed from February. Write $\widehat\lambda_1\ge\cdots\ge\widehat\lambda_7$ for its eigenvalues. The exact SAA solution is the leading two-dimensional eigenspace, represented by $[\widehat{\mathbf X}_N]$.

For an orthonormal complement $\mathbf X_\perp\in\R^{7\times5}$, a horizontal tangent vector has the form $\mathbf X_\perp\mathbf K$, with $\mathbf K\in\R^{5\times2}$. The ten-dimensional score is $-2\operatorname{vec}(\mathbf X_\perp^\top zz^\top\mathbf X)$, where $\operatorname{vec}$ stacks columns. At the empirical minimizer, choosing both frames from the eigenvectors of $\mathbf C_N$ makes the Hessian diagonal with entries $2(\widehat\lambda_j-\widehat\lambda_k)$ for $j\le2<k$. Its smallest eigenvalue is therefore $2(\widehat\lambda_2-\widehat\lambda_3)$, linking the empirical eigengap to local curvature. This Hessian and the long-run tangent-score covariance determine the sandwich uncertainty.

We estimate the long-run score covariance using nonoverlapping batches of 180 readings (nominally 30 minutes), restarting after timestamp gaps longer than 30 seconds. Complete batches retain 98.0\% of evaluation readings. The second moment, empirical Hessian, and denominator $N$ in the root-mean-square (RMS) subspace uncertainty $\{\tr(\wh{\mathsf V}_N^\dagger)/N\}^{1/2}$ use the full window. Evaluating the sandwich covariance at exact SAA or at the one-step estimate gives identical RMS values to Table~\ref{tab:metropt} precision. Batch lengths of 60 and 360 readings provide sensitivity comparisons.

To compare numerical error with this statistical uncertainty, we initialize Riemannian gradient descent at the February reference subspace, use the Grassmann exponential with step size $1/(4\widehat\lambda_1)$, and stop when the gradient norm is at most $0.1$. At $N\approx50{,}000$, this fixed tolerance is comparable in magnitude to the one-step pilot scale $N^{-1/4}\approx0.07$, whereas the direct-approximation scale is $N^{-1/2}\approx0.0045$. The solver takes 8 to 37 iterations, and all seven pilot Hessians are positive definite.

We then apply one exact intrinsic Newton correction at each pilot. Put $\mathbf A=\mathbf X^\top\mathbf C_N\mathbf X$, $\mathbf M=\mathbf X_\perp^\top\mathbf C_N\mathbf X_\perp$, and $\mathbf B=\mathbf X_\perp^\top\mathbf C_N\mathbf X$. Solve the Sylvester equation $\mathbf K\mathbf A-\mathbf M\mathbf K=\mathbf B$ and take $\Exp_{\mathbf X}(\mathbf X_\perp\mathbf K)$ as the one-step estimate. Because the Grassmann trace objective also admits an exact solution by eigendecomposition, distances from both the pilot and its correction to $[\widehat{\mathbf X}_N]$ isolate numerical optimization error on the observed trajectory. Comparing these distances with RMS uncertainty shows whether solver error is material on the estimated sampling scale.

\subsection{Dependence, numerical accuracy, and operating changes}

Temporal dependence substantially increases estimated subspace uncertainty. At batch size 180, the sandwich trace is 8.9 to 42.2 times its i.i.d.\ counterpart, corresponding to a 3.0- to 6.5-fold increase in RMS uncertainty (Figure~\ref{fig:metropt}a). For batch sizes 60 and 360, the corresponding trace ratios range from 11.6 to 27.4 and from 10.1 to 71.2. Together with lag-one score correlations reaching 0.98, this variation shows how the estimated uncertainty responds to persistent dependence across compressor operating cycles.

\begin{table}[!htb]
\centering
\caption{Summary of the seven chronological MetroPT-3 evaluation windows in 2020. Load is the percentage of readings for which the electric valve is active, $d_{\rm Feb}$ is the geodesic distance (in radians) from the February reference subspace, and RMS is the uncertainty scale defined in Section~\ref{sec:metropt-computation}.}
\label{tab:metropt}
\small
\setlength{\tabcolsep}{9pt}
\begin{tabular}{cccccc}
\toprule
Start & $N$ & Eigengap & Load (\%) & $d_{\rm Feb}$ & RMS \\
\midrule
03-02 & 51,497 & 3.30 & 15.6 & 0.809 & 0.033 \\
03-09 & 48,442 & 3.03 & 21.4 & 0.861 & 0.044 \\
03-16 & 55,339 & 2.24 & 8.0 & 0.482 & 0.018 \\
03-23 & 53,283 & 2.24 & 26.8 & 0.970 & 0.030 \\
03-30 & 47,078 & 2.60 & 10.7 & 0.608 & 0.014 \\
04-06 & 52,221 & 2.38 & 17.7 & 0.773 & 0.039 \\
04-13 & 32,591 & 2.09 & 7.3 & 0.435 & 0.034 \\
\bottomrule
\end{tabular}
\end{table}

Table~\ref{tab:metropt} shows positive empirical curvature in all seven windows: the eigengap $\widehat\lambda_2-\widehat\lambda_3$ ranges from 2.09 to 3.30, so the leading empirical subspace is locally well identified. It explains 85.8\% to 90.3\% of the transformed second moment. The associated within-window RMS uncertainty ranges from 0.014 to 0.044 radians, providing the reference scale for both numerical accuracy and operating changes.

Subspace rotation relative to February generally increases with compressor load. The two lightest-load windows, at 7.3\% and 8.0\%, have distances $d_{\rm Feb}=0.435$ and $0.482$ radians, whereas the heaviest-load window, at 26.8\%, has the largest distance, $0.970$ radians. The ordering is not strictly monotone across the intermediate windows, but the overall association connects changes in the geometric operating summary with duty-cycle variation. These cross-window shifts exceed the within-window RMS scales, motivating window-specific estimation.

The coarsely stopped pilots lie 0.010 to 0.019 radians from the exact empirical subspaces, or 0.36 to 1.02 times the estimated RMS uncertainty. One intrinsic Newton correction reduces these discrepancies by roughly three to four orders of magnitude, to between $1.4\times10^{-6}$ and $1.0\times10^{-5}$ radians, at most $3.4\times10^{-4}$ times the RMS scale (Figure~\ref{fig:metropt}b). Thus a single correction makes numerical optimization error negligible relative to the estimated sampling uncertainty in every window. The application illustrates both the sensitivity of subspace uncertainty to serial dependence and the value of calibrating solver accuracy against a statistical scale.

\section{Discussion}
\label{sec:discussion}


A computed estimator supports valid statistical inference whenever its numerical error is negligible on the scale of sampling fluctuations. The population loss Hessian and the long-run score covariance determine that sampling scale. We estimate it from a single trajectory using the empirical Hessian and batch means of scores evaluated in the same tangent space. On the fixed model class in Theorem~\ref{thm:general_manifold_minimax}, exact SAA attains the minimax rate. Pairing a coarse pilot iterate with a single intrinsic inexact Newton step attains exact-SAA first-order efficiency under substantially relaxed optimization tolerances. In the MetroPT-3 compressor analysis, a single Newton step reduces numerical discrepancies from the exact empirical subspaces to orders of magnitude below the estimated sampling uncertainty across all seven windows.

{Several directions merit further study. On the probabilistic side, while Brownian strong invariance underpins Assumption~\ref{ass:batch_means}, establishing nonasymptotic Berry-sseen coverage rates and data-driven optimal batch-length rules under pure $L^2$ spectral contraction, as well as relaxing uniform boundedness to heavy-tailed losses and nonstationary initializations, remains an important problem. Geometrically, allowing the intrinsic dimension $m$ to grow with $N$,where both the entropy integral $\cJ_\M$ and the batch-rank requirement $a_N\ge m+1$ become active, and developing adaptive localization procedures would broaden applicability.

\section*{Acknowledgments}
 Qirui Hu's research was supported by National Natural Science Foundation of China (NSFC) (Grant Nos.12601520), the Shanghai Engineering Research Center of Finance Intelligence (Grant No.~19DZ2254600) and by TRR 391 \textit{Spatio-temporal Statistics for the Transition of Energy and Transport} (Project number 520388526) funded by the Deutsche Forschungsgemeinschaft (DFG, German Research Foundation).

\clearpage
\appendix
\numberwithin{figure}{section}
\numberwithin{table}{section}

\section*{Supplementary Material}
%
%

\section{Supporting results for estimation and inference}
\label{sec:supporting-results}
\subsection{Uniform concentration}
\label{sec:uniform-details}
Compactness and continuity give a nonempty compact empirical argmin. Its graph is Borel because $\min_x L_N(x)$ is the infimum over a countable dense set. The Arsenin--Kunugui theorem \citep[Theorem~18.18]{kechris1995} gives a measurable exact minimizer.


Define $\Delta_N=\sup_{x\in\M}|L_N(x)-f(x)|$. The entropy integral in \eqref{eq:concentration_radii} is finite: a finite cover by interior coordinate balls and boundary half-balls with bounded metric coefficients gives $\cN(\M,d,\eps)=O(\eps^{-m})$ as $\eps\downarrow0$.



\begin{theorem}
\label{thm:uniform_saa}
Under Assumptions~\ref{ass:global} and~\ref{ass:spectral}, the radius in \eqref{eq:concentration_radii} satisfies, for every $\delta\in(0,1)$, with probability at least $1-\delta$,
\begin{equation}
  \Delta_N\le\mathfrak D_N(\delta).
  \label{eq:uniform_chaining_bound}
\end{equation}
If, for $0<\eps\le D_\M$,
\begin{equation}
  \cN(\M,d,\eps)\le(A_\M/\eps)^m,
  \qquad A_\M\ge D_\M,
  \label{eq:polynomial_entropy}
\end{equation}
then
\begin{equation}
  \cJ_\M
  \le D_\M\left[
    \sqrt{\log 2+m\log(A_\M/D_\M)}
    +\frac{\sqrt{\pi m}}{2}
  \right].
  \label{eq:entropy_integral_manifold}
\end{equation}
Consequently, for fixed bounded geometry,
\begin{equation*}
  \Delta_N
  =O_{\Pbb}\!\left(
    (B+GD_\M)\sqrt{\frac{\tau_\lambda m}{N}}
  \right),
\end{equation*}
with a sample-size scaling free of logarithmic factors.
\end{theorem}


\subsection{Complete local bounds}
\label{sec:local-details}


The following results supplement the parameter bounds in Theorem~\ref{thm:local_fast_rate}.



\begin{theorem}[Additional local bounds]
\label{thm:local_fast_rate_full}
Under the assumptions and confidence conditions of Theorem~\ref{thm:local_fast_rate}, its common event can be chosen so that $\Delta_N\le\mathfrak D_N(\delta_0)$, $\|\grad L_N(x^*)\|\le s_N$, and $\|\Hess L_N(x^*)-\mathsf H\|_\op\le h_N$ hold simultaneously. On this event,
\begin{equation}
 \Hess L_N(x)\succeq(\lambda_*/2)I,\qquad x\in U_{r/2}.
 \label{eq:intrinsic_empirical_strong_convexity}
\end{equation}
Every objective-gap solution covered by that theorem also satisfies
\begin{equation}
 f(\wt x_N)-f(x^*)\le\frac{K_2}{2}
 \left(\frac{2s_N}{\lambda_*}+2\sqrt{\frac{\eps_N^{\rm obj}}{\lambda_*}}\right)^2.
 \label{eq:intrinsic_fast_excess_risk}
\end{equation}
For a localized gradient-residual solution covered by that theorem,
\begin{equation}
 d(\wt x_N^{\rm grad},\wh x_N)\le\frac{2\eps_N^{\rm grad}}{\lambda_*}.
 \label{eq:intrinsic_gradient_distance}
\end{equation}
\end{theorem}



The Hessian requirement $h_N\le\lambda_*/4$ is satisfied if $N\ge400\tau_\lambda K_2^2\{m+\log(6/\delta)\}/\lambda_*^2$; the localization requirement remains as stated in Theorem~\ref{thm:local_fast_rate}.



\begin{theorem}[Finite-sample Bahadur representation]
\label{thm:bahadur_full}
Under Assumptions~\ref{ass:global}--\ref{ass:local}, consider the estimators and the joint event for localization, score, and Hessian bounds specified in Theorem~\ref{thm:local_fast_rate_full}. Define the common remainder radius
\[
 \mathfrak r_N^{\rm B}(\delta)
 :=\frac{2h_Ns_N}{\lambda_*^2}+\frac{2K_3s_N^2}{\lambda_*^3}.
\]
For the exact SAA estimator,
\begin{equation}
\norm{\Log_{x^*}(\wh x_N)+\mathsf H^{-1}g_N}
  \le
  \mathfrak r_N^{\rm B}(\delta).
  \label{eq:bahadur_exact_intrinsic}
\end{equation}
For the approximate-estimator bounds, let $C_{\Log}<\infty$
satisfy
\[
  \norm{\Log_{x^*}(x)-\Log_{x^*}(y)}\le C_{\Log}d(x,y)
\]
for all $x,y\in U_r$. Such a constant exists because the smooth normal-coordinate inverse has bounded differential on $\overline U_r$ and $U_r$ is strongly geodesically convex.
For an $\eps_N^{\rm obj}$-approximate solution $\wt x_N$,
\begin{equation*}
  \norm{\Log_{x^*}(\wt x_N)+\mathsf{H}^{-1}g_N}
  \le
  \mathfrak r_N^{\rm B}(\delta)
  +2C_{\Log}\sqrt{\frac{\eps_N^{\rm obj}}{\lambda_*}}.
\end{equation*}
When $2\mathfrak D_N(\delta_0)<\gamma_0$ and $h_N\le\lambda_*/4$, any estimator $\wt x_N^{\rm grad}\in U_{r/2}$ satisfying the gradient-residual bound \eqref{eq:gradient_certificate} obeys
\begin{equation}
  \norm{\Log_{x^*}(\wt x_N^{\rm grad})+\mathsf{H}^{-1}g_N}
  \le
  \mathfrak r_N^{\rm B}(\delta)
  +\frac{2C_{\Log}\eps_N^{\rm grad}}{\lambda_*}.
  \label{eq:bahadur_gradient_intrinsic_first}
\end{equation}
\end{theorem}


At fixed confidence and with fixed local constants, the exact-estimator bound in \eqref{eq:bahadur_exact_intrinsic} is of order $\tau_\lambda m/N$ whenever the localization and Hessian concentration conditions hold. This is the remainder measured in the torus experiment.

\subsection{Statistical one-step expansion}
Choose constants $\rho_{\Exp}\in(0,r/2)$ and $C_{\Exp}<\infty$ so that at every $x\in\overline U_{r/2}$ the exponential map is feasible, injective, and $C_{\Exp}$-Lipschitz on the closed tangent ball of radius $\rho_{\Exp}$, with image in $U_r$.

\begin{theorem}[Deterministic one-step contraction]
\label{thm:one_step_quadratic}
Under the local geometry of Assumption~\ref{ass:local}, fix a realization of $L_N$ twice continuously differentiable on $U_r$ satisfying \eqref{eq:hessian_transport_lipschitz} with constant $K_3$ and $\Hess L_N(x)\succeq(\lambda_*/2)I$ on $U_{r/2}$. Suppose $\wh x_N,\bar x_N\in U_{r/2}$ with $\grad L_N(\wh x_N)=0$, set $d_N=d(\bar x_N,\wh x_N)$, and let $v_N^{\rm os}\in T_{\bar x_N}\M$ satisfy \eqref{eq:one_step_direction_final}. If
\begin{equation*}
  d_N+\frac{K_3}{\lambda_*}d_N^2+\frac{2\eta_N}{\lambda_*}\le\rho_{\Exp},
\end{equation*}
then $v_N^{\rm os}\in D_{\bar x_N}$, $x_N^{\rm os}=\Exp_{\bar x_N}(v_N^{\rm os})\in U_r$, and
\begin{equation}
  d(x_N^{\rm os},\wh x_N)
  \le C_{\Exp}\left(
    \frac{K_3}{\lambda_*}d_N^2
    +\frac{2\eta_N}{\lambda_*}
  \right).
  \label{eq:one_step_quadratic_bound}
\end{equation}
Under the same step-radius condition, for an objective-gap tolerance $\bar\eps_N^{\rm obj}\ge0$, a pilot $\bar x_N\in U_{r/2}$ satisfying $L_N(\bar x_N)\le L_N(\wh x_N)+\bar\eps_N^{\rm obj}$ obeys
\begin{equation}
  d(x_N^{\rm os},\wh x_N)
  \le
  \frac{4C_{\Exp}K_3}{\lambda_*^2}\bar\eps_N^{\rm obj}
  +\frac{2C_{\Exp}}{\lambda_*}\eta_N.
  \label{eq:one_step_gap_bound}
\end{equation}
For a gradient-residual tolerance $\bar\eps_N^{\rm grad}\ge0$, a pilot $\bar x_N\in U_{r/2}$ satisfying $\norm{\grad L_N(\bar x_N)}\le\bar\eps_N^{\rm grad}$ obeys
\begin{equation}
  d(x_N^{\rm os},\wh x_N)
  \le
  \frac{4C_{\Exp}K_3}{\lambda_*^3}(\bar\eps_N^{\rm grad})^2
  +\frac{2C_{\Exp}}{\lambda_*}\eta_N.
  \label{eq:one_step_gradient_bound}
\end{equation}
\end{theorem}

On the joint event of Theorem~\ref{thm:local_fast_rate_full}, the logarithmic displacement of the corrected estimate has the same linear term $-\mathsf H^{-1}g_N$ as exact SAA. Under the step-radius condition, its remainder is bounded by the right side of \eqref{eq:bahadur_exact_intrinsic} plus $C_{\Log}$ times the right side of \eqref{eq:one_step_quadratic_bound}, by the triangle inequality on $U_r$.

\subsection{Expected risk of exact SAA}


\begin{theorem}
\label{thm:expected_risk}
Under Assumptions~\ref{ass:global}--\ref{ass:local}, let $\wh x_N$ be an exact SAA estimator and set $\sigma_*^2=\tr(\mathsf C_0)$. The finite-trajectory covariance from Section~\ref{sec:local} has the representation
\begin{equation*}
 \mathsf S_N=\mathsf C_0+
 \sum_{h=1}^{N-1}\left(1-\frac hN\right)(\mathsf C_h+\mathsf C_h^*).
\end{equation*}
The contraction also gives the operator inequality
\begin{equation}
  0\preceq\mathsf{S}_N\preceq\tau_\lambda\mathsf{C}_0.
  \label{eq:finite_covariance_envelope}
\end{equation}
Then
\begin{equation*}
  \E\norm{\mathsf{H}^{-1}g_N}^2
  =\frac1N\tr(\mathsf{H}^{-1}\mathsf{S}_N\mathsf{H}^{-1})
  \le\frac{\tau_\lambda}{N}
       \tr(\mathsf{H}^{-1}\mathsf{C}_0\mathsf{H}^{-1}).
\end{equation*}
Suppose the exact-estimator conditions of \cref{thm:local_fast_rate_full} hold at confidence level $\delta$, and set $\delta_0=\delta/3$. Then
\begin{align}
  \E d(\wh x_N,x^*)^2
  &\le
  \frac{4\tau_\lambda\sigma_*^2}{\lambda_*^2N}
  +D_\M^2\delta,
  \label{eq:expected_squared_parameter}\\
  \E\{f(\wh x_N)-f(x^*)\}
  &\le
  \frac{2K_2\tau_\lambda\sigma_*^2}{\lambda_*^2N}
  +2B\delta.
  \label{eq:expected_excess_risk}
\end{align}
More sharply, use the remainder radius $\mathfrak r_N^{\rm B}(\delta)$ from Theorem~\ref{thm:bahadur_full}. 
Set $T_N=\tr(\mathsf{H}^{-1}\mathsf{S}_N\mathsf{H}^{-1})$. Then
\begin{align}
 \left|\E d(\wh x_N,x^*)^2-\frac{T_N}{N}\right|
 &\le2\mathfrak r_N^{\rm B}(\delta)\sqrt{\frac{T_N}{N}}+\{\mathfrak r_N^{\rm B}(\delta)\}^2
   +\left(D_\M^2+\frac{K_1^2}{\lambda_*^2}\right)\delta,
 \label{eq:hessian_adapted_expected_parameter}\\
 \left|\E\{f(\wh x_N)-f(x^*)\}
        -\frac{\tr(\mathsf{H}^{-1}\mathsf{S}_N)}{2N}\right|
 &\le\mathcal R_N^f(\delta)
       +\left(2B+\frac{K_1^2}{2\lambda_*}\right)\delta.
 \label{eq:hessian_adapted_expected_risk}
\end{align}
Here the explicit remainder is
\begin{equation}
 \mathcal R_N^{f}(\delta)
 :=\mathfrak r_N^{\rm B}(\delta)
       \sqrt{\frac{\tr(\mathsf{S}_N)}{N}}
   +\frac{K_2}{2}\{\mathfrak r_N^{\rm B}(\delta)\}^2
   +\frac{4K_3s_N(\delta_0)}{3\lambda_*^3N}\tr(\mathsf{S}_N).
 \label{eq:excess_remainder}
\end{equation}
In particular, the leading excess-risk term is bounded by
$\tau_\lambda\tr(\mathsf{H}^{-1}\mathsf{C}_0)/(2N)$.
\end{theorem}



\begin{corollary}
\label{cor:unconditional_expected_risk}
Under Assumptions~\ref{ass:global}--\ref{ass:local}, with fixed dimension, contraction, and local constants, the exact-estimator conditions of Theorem~\ref{thm:local_fast_rate} hold at $\delta_N=N^{-2}$ for all sufficiently large $N$. Moreover,
\begin{equation}
 \mathfrak r_N^{\rm B}(N^{-2})=O((\log N)/N),
 \qquad \mathcal R_N^f(N^{-2})=O((\log N)/N^{3/2}).
 \label{eq:unconditional_bahadur_order}
\end{equation}
\end{corollary}



\subsection{Strong approximation and batch growth}
\label{sec:plugin-details}



\begin{proposition}[Strong approximation under geometric ergodicity]
\label{prop:checkable_sip}
Suppose the stationary chain is $\pi$-Harris recurrent and geometrically ergodic in total variation. Let $\zeta$ be a bounded, mean-zero score in the fixed $m$-dimensional space $V$, with long-run covariance $\mathsf S$. For every $\rho\in(0,1/4)$, on a suitable probability space extension there exist a standard $\R^m$-valued Brownian motion $W$, a linear map $\mathsf L_S:\R^m\to V$ with $\mathsf L_S\mathsf L_S^*=\mathsf S$, and an almost surely finite random variable $C$ such that, eventually almost surely,
\begin{equation}
  \max_{1\le k\le N}
  \norm{\sum_{i=1}^k\zeta(\xi_i)-\mathsf L_S W(k)}
  \le C N^{1/2-\rho}.
  \label{eq:score_sip_final}
\end{equation}
Given any fixed $\nu\in(1/2,1)$, choose $\rho$ so that
\begin{equation}
  (1-\nu)/2<\rho<1/4.
  \label{eq:polynomial_batch_rule}
\end{equation}
The batch sizes in Assumption~\ref{ass:batch_means} then satisfy
\begin{equation}
  b_N\to\infty,\qquad a_N\to\infty,\qquad N^{1-2\rho}/b_N\to0.
  \label{eq:batch_growth_final}
\end{equation}
\end{proposition}


\FloatBarrier
\section{Minimax constructions}
\label{sec:minimax-constructions}
\subsection{The product torus}
\label{sec:torus-construction}

Fix dimension $m\ge1$, scale $a\in(0,1]$, and $L^2$ contraction parameter $\lambda\in[0,1)$ independently of
sample size. The parameter class is the fixed cube
$\Theta_m=[-1/4,1/4]^m$. For $\theta\in\Theta_m$, define on
$\Xi=\{-1,+1\}^m$ the product distribution
\begin{equation}
 \pi_\theta(z)=\prod_{j=1}^m\frac{1+\theta_jz_j}{2}.
 \label{eq:hypercube_stationary_law}
\end{equation}
The stationary refresh kernel on $\{-1,+1\}^m$ is
\begin{equation}
 \mathsf P_\theta(z,\cdot)=\lambda\delta_z+(1-\lambda)\pi_\theta,
 \label{eq:refresh_chain}
\end{equation}
where $\delta_z$ denotes the Dirac measure at $z$.
It is reversible and its mean-zero $L^2(\pi_\theta)$ contraction
coefficient equals $\lambda$ for every $\theta$. Let $Z_1,\ldots,Z_N$ denote a stationary trajectory from this kernel, with path law $\Pbb_\theta^{(N)}$ and expectation $\E_\theta$. The testing parameters have the form $\theta=\delta\omega$, with a sign vector $\omega\in\{-1,+1\}^m$ and $0<\delta\le1/4$.

Write $x=(x_1,\ldots,x_m)\in\Torus^m=(\mathbb S^1)^m$ with
$x_j\in\mathbb S^1\subset\R^2$, equip the product with its flat product metric, and use the common loss
\begin{equation}
  \ell(x,z)
  =\frac1m\sum_{j=1}^m
    \left[1-\inner{x_j}{(1,az_j)}_{\R^2}\right].
  \label{eq:torus_loss}
\end{equation}
The loss is smooth and bounded independently of $m$. Under
$\pi_\theta$, its unique population minimizer has coordinates
\begin{equation*}
  x_{\theta,j}^* =\frac{(1,a\theta_j)}{\sqrt{1+a^2\theta_j^2}}.
\end{equation*}
Put $\kappa_N=1+(N-1)(1-\lambda)$, the expected number of independent draws including the initial state. The lower-bound construction uses $\delta_N=1/(4\sqrt{\kappa_N})$.
The finite testing subset $\{\delta_N\omega:\omega\in\{-1,+1\}^m\}$
lies inside the fixed cube $\Theta_m$ for every
$N\ge1$.

Write $\overline Z_N=N^{-1}\sum_{i=1}^N Z_i$, with coordinates $\overline Z_{N,j}$. The unique empirical
minimizer has coordinates
\begin{equation}
  \wh x_{N,j}^{\rm tor}
  =\frac{(1,a\overline Z_{N,j})}{\sqrt{1+a^2\overline Z_{N,j}^2}},
  \qquad j=1,\ldots,m.
  \label{eq:exact_torus_saa}
\end{equation}
Define the finite-trajectory variance factor
\begin{equation*}
 v_N(\lambda)=\frac{N+2\sum_{h=1}^{N-1}(N-h)\lambda^h}{N^2}.
\end{equation*}

\begin{proposition}
\label{prop:torus_upper}
For every $N\ge1$ and $\theta\in\Theta_m$, exact SAA satisfies
\begin{align}
 \E_\theta d_{\Torus^m}(\wh x_N^{\rm tor},x_\theta^*)^2
 &\le a^2v_N(\lambda)\sum_{j=1}^m(1-\theta_j^2),
 \label{eq:torus_upper_exact_variance}\\
 \sup_{\theta\in\Theta_m}\E_\theta
 d_{\Torus^m}(\wh x_N^{\rm tor},x_\theta^*)^2
 &\le a^2m\min\{1,\tau_\lambda/N,2/\kappa_N\}.
 \label{eq:torus_upper_uniform}
\end{align}
\end{proposition}

On the testing subfamily $\theta=\delta\omega$, the population Hessian matrix is $\mathsf{H}=\sqrt{1+a^2\delta^2}I_m/m$, and
$\tr(\mathsf{H}^{-1}\mathsf{C}_0\mathsf{H}^{-1})=a^2m(1-\delta^2)/(1+a^2\delta^2)^2$;
this matches the information scale in the upper bound.

Define
\[
 \mathfrak R_{N,m}(a,\lambda)
 =\inf_{\wt x_N}\sup_{\theta\in\Theta_m}
 \E_\theta d_{\Torus^m}(\wt x_N,x_\theta^*)^2,
\]
where the infimum ranges over all estimators based on the full trajectory.

\begin{proposition}
\label{prop:torus_minimax_constants}
For every $m,N\ge1$, $a\in(0,1]$, and $\lambda\in[0,1)$,
\begin{equation}
 \frac{a^2m}{256\kappa_N}
 \le \mathfrak R_{N,m}(a,\lambda)
 \le \sup_{\theta\in\Theta_m}\E_\theta
 d_{\Torus^m}(\wh x_N^{\rm tor},x_\theta^*)^2
 \le \frac{2a^2m}{\kappa_N}.
 \label{eq:minimax_lower_exact}
\end{equation}
If $N(1-\lambda)\ge1$, this implies
\begin{equation}
 \frac{a^2m}{1024}\frac{\tau_\lambda}{N}
 \le \mathfrak R_{N,m}(a,\lambda)
 \le a^2m\frac{\tau_\lambda}{N}.
 \label{eq:minimax_tau_order}
\end{equation}
\end{proposition}


For the $1/m$-normalized loss \eqref{eq:torus_loss}, let $\mathfrak E_{N,m}$ be the corresponding minimax expected excess risk over the same estimators and parameters. If $t_j\in[-\pi,\pi]$ is the shortest angular displacement from $x_{\theta,j}^*$, then
\[
 f_\theta(x)-f_\theta(x_\theta^*)=
 m^{-1}\sum_j\sqrt{1+a^2\theta_j^2}\,(1-\cos t_j).
\]
Since $2t_j^2/\pi^2\le1-\cos t_j\le t_j^2/2$ and $1\le\sqrt{1+a^2\theta_j^2}\le\sqrt{17}/4$, \eqref{eq:minimax_lower_exact} gives
\[
 \frac{a^2}{128\pi^2\kappa_N}\le\mathfrak E_{N,m}
 \le\sup_\theta\E_\theta\{f_\theta(\wh x_N^{\rm tor})-f_\theta(x_\theta^*)\}
 \le\frac{\sqrt{17}\,a^2}{4\kappa_N}.
\]
The lower comparison applies to every estimator; the upper comparison uses exact SAA. When $N(1-\lambda)\ge1$, \eqref{eq:minimax_tau_order} gives excess-risk order $a^2\tau_\lambda/N$.


\subsection{A local family on a compact manifold}
\label{sec:general_manifold_minimax}


We now give the local construction and explicit constants for Theorem~\ref{thm:general_manifold_minimax}.


Fix $x_0\in\operatorname{int}\M$. Choose $r>0$ small enough that $\overline U_{2r}$ lies in a strongly convex normal neighborhood in the interior, where
\begin{equation*}
  \Exp_{x_0}:B_{2r}(0)\subset T_{x_0}\M\longrightarrow U_{2r}\subset\operatorname{int}\M
\end{equation*}
is a diffeomorphism and, for some constants
$0<c_-\le c_+<\infty$,
\begin{equation}
  c_-\norm{u-v}
  \le
  d\!\left(\Exp_{x_0}u,\Exp_{x_0}v\right)
  \le
  c_+\norm{u-v},
  \qquad u,v\in B_r(0).
  \label{eq:general_chart_bilip}
\end{equation}
For a fixed smooth manifold one may always shrink $r$ so that, for example,
$c_-=1/2$ and $c_+=2$.

Choose a smooth convex nondecreasing function
$\eta:[0,\infty)\to[0,\infty)$ such that
\begin{equation}
  \eta(s)=0\quad\text{for }s\le\frac1{16},
  \qquad
  \eta'(s)\ge1\quad\text{for }s\ge1.
  \label{eq:barrier_eta}
\end{equation}
Define
\begin{equation}
  \Psi_{m,r}(u)
  :=\frac12\norm{u}^2
    +\sqrt m\,r^2
     \eta\!\left(\frac{\norm{u}^2}{r^2}\right),
  \qquad u\in\R^m,
  \label{eq:chart_potential}
\end{equation}
and
\begin{equation}
  T_r(y)
  :=\argmin_{u\in\R^m}
  \left\{\Psi_{m,r}(u)-\inner{y}{u}\right\}.
  \label{eq:chart_response_map}
\end{equation}
We identify $T_{x_0}\M$ with $\R^m$ through an orthonormal frame.


The minimizer $T_r$ is unique
and one-Lipschitz, maps every $y$ with $\norm y\le r\sqrt m$ into $B_r(0)$,
and equals the identity when $\norm y\le r/4$.


Let $\chi\in C^\infty(\M;[0,1])$ equal one on $U_r$ and have support
in $U_{2r}$. For $z\in\{-1,+1\}^m$, define the common loss
\begin{equation}
  \ell_{m,r}^{\rm ch}(x,z)
  :=C+\chi(x)\left[
       \Psi_{m,r}(\Log_{x_0}x)
       -r\inner{z}{\Log_{x_0}x}
       -C
     \right],
  \label{eq:chart_loss}
\end{equation}
where $C>0$ and the bracketed smooth function is used only on
$\operatorname{supp}(\chi)\subset U_{2r}$. Thus
\eqref{eq:chart_loss} extends to a bounded smooth function on all of
$\M\times\{-1,+1\}^m$.

For every observed trajectory, the unique empirical minimizer is
\begin{equation}
  \wh x_N^{\rm ch}=\Exp_{x_0}T_r(r\overline Z_N).
  \label{eq:chart_exact_saa}
\end{equation}


Use the parameter cube $\Theta_m$ and product laws $\pi_\theta$ from Appendix~\ref{sec:torus-construction}. For each
$\theta$ let
\begin{equation}
 \mathfrak K_\theta(\lambda)
 :=\{\mathsf P:\pi_\theta\mathsf P=\pi_\theta,\quad
 \|\mathsf P-\Pi_\theta\|_{L_0^2(\pi_\theta)\to L_0^2(\pi_\theta)}
 \le\lambda\}.
 \label{eq:general_spectral_kernel_class}
\end{equation}
Every trajectory starts in stationarity, and its target is $x_\theta^*=\Exp_{x_0}T_r(r\theta)$.


Define the fixed-class risk
\[
 \mathfrak R_{N,m}^{\M}(r,\lambda)
 :=\inf_{\wt x_N}\sup_{\theta\in\Theta_m}
       \sup_{\mathsf P\in\mathfrak K_\theta(\lambda)}
       \E_{\theta,\mathsf P}d(\wt x_N,x_\theta^*)^2.
\]
The infimum ranges over all estimators based on the full trajectory. For the lower-bound test, set
\begin{equation}
 \delta_{N,m}=\frac1{4\sqrt{\kappa_N\vee m}},\qquad
 \theta=\delta_{N,m}\omega.
 \label{eq:general_signal_level}
\end{equation}
These vertices lie in $\Theta_m$ and satisfy
$T_r(r\delta_{N,m}\omega)=r\delta_{N,m}\omega$.


\begin{proposition}
\label{prop:general_manifold_minimax_full}
Fix $m\ge1$, $r,c_-,c_+$ as above, and $\lambda\in[0,1)$.
For every $N\ge1$, the loss \eqref{eq:chart_loss} and fixed class
$\theta\in\Theta_m$, $\mathsf P\in\mathfrak K_\theta(\lambda)$ satisfy
\begin{equation}
 \begin{aligned}
  \frac{c_-^2r^2}{64}\frac{m}{\kappa_N\vee m}
  &\le
  \mathfrak R_{N,m}^{\M}(r,\lambda)
  \\
  &\le
  \sup_{\theta,\mathsf P}
  \E_{\theta,\mathsf P}
  d(\wh x_N^{\rm ch},x_\theta^*)^2
  \le
  4c_+^2r^2\frac{m}{\kappa_N\vee m}.
 \end{aligned}
 \label{eq:general_minimax_kappa_sandwich}
\end{equation}
Moreover,
\begin{equation}
 \begin{aligned}
  \frac{c_-^2r^2}{256}
  \min\left\{1,\frac{m\tau_\lambda}{N}\right\}
  &\le
  \mathfrak R_{N,m}^{\M}(r,\lambda)
  \\
  &\le
  4c_+^2r^2
  \min\left\{1,\frac{m\tau_\lambda}{N}\right\}.
 \end{aligned}
 \label{eq:general_minimax_tau_sandwich}
\end{equation}
\end{proposition}


The chart constants in \eqref{eq:general_minimax_kappa_sandwich} are uniform along a sequence of manifolds if \eqref{eq:general_chart_bilip} holds with common $r,c_-,c_+$. The chart targets have squared diameter at most $4c_+^2r^2$, whereas the global torus cube in Section~\ref{sec:torus-construction} can have squared diameter of order $a^2m$.

\FloatBarrier
\section{Additional Experiments}
\label{sec:additional-experiments}

\subsection{Simulation design}
\label{sec:simulation-design}


For the rate, Bahadur remainder, and one-step experiments in Section~\ref{sec:numerical-additional}, we generate 1000 independent trajectories per setting and compare estimators on the same trajectory. We use the whole-vector refresh kernel \eqref{eq:refresh_chain} with product law \eqref{eq:hypercube_stationary_law}. The sphere risk comparison also uses independently generated trajectories from the coordinatewise refresh kernel with the same contraction. This kernel independently retains each coordinate with probability $\lambda$ and otherwise redraws it from its stationary marginal.



For the uniform objective deviation on $\mathbb S^m$, the chart radius is $r=0.35$ and the signs are centered. We cross $m\in\{2,5,10,20,50\}$, $\lambda\in\{0,0.6,0.8,0.9\}$, and $N=2^k$ for $k=7,\ldots,13$. We normalize the mean deviation by multiplying by $\sqrt{N/(\tau_\lambda m)}$. For sphere risk, we use the signal level in \eqref{eq:general_signal_level}, $m\in\{5,20,100\}$, $N\in\{256,1024,4096\}$, and $\lambda\in\{0,0.8,0.95\}$ in \eqref{eq:chart_exact_saa}. The torus risk design uses the same $N$ and $\lambda$, with $m\in\{5,20,50\}$, $a=1$, and the testing signal $\delta_N$ from Appendix~\ref{sec:torus-construction} in \eqref{eq:torus_loss}. Its two reference scales are $a^2m/\kappa_N$ and $a^2m(1-\delta_N^2)v_N(\lambda)$. For the torus Bahadur remainder, we use $\delta=0.15$, $m\in\{2,5,10\}$, and the $N$ and $\lambda$ grids of the uniform-deviation experiment.



The one-step experiment uses a modified torus loss in angular coordinates,
\begin{equation}
 \ell(\theta,z)=\frac1m\sum_{j=1}^m
 \{1-\cos\theta_j+b(1-\cos 2\theta_j)-az_j\sin\theta_j\},
 \label{eq:simulation-asymmetric-loss}
\end{equation}
with $m=3$, $a=0.5$, $b=0.05$, common sign mean $\delta=0.2$, and $\lambda=0.8$.
At each of the seven powers of two above, we perturb the empirical minimizer in a random unit direction until the objective gap or gradient norm reaches $0.05N^{-p}$. The objective-gap exponents are $0.4,0.6,0.8,1.2$, and the gradient exponents are $0.2,0.3,0.4,0.6$. For the inexact solve, we use an objective-gap pilot of order $N^{-0.8}$ and a linear-system residual $0.05N^{-q}$ with $q\in\{0.4,0.6,1\}$.



For coverage, we use two models. The flat model is the torus loss
\eqref{eq:torus_loss} with $m=2$, $a=1$, and sign mean $\delta=0.15$.
The curved model has decision space $\mathbb S^2$ with its round metric and
loss $\ell(x,z)=1-x^\top z$, where
$z=(0.6Z_1,0.4Z_2,1)^\top$ and each sign has mean $0.15$.
The whole-vector refresh kernel acts on the signs. With $\bar z_N=N^{-1}\sum_{i=1}^Nz_i$ and population mean $\mu=(0.09,0.06,1)^\top$, the population direction is
$x^*=\mu/\|\mu\|$ and exact SAA is $\wh x_N=\bar z_N/\|\bar z_N\|$.
For an orthonormal tangent frame $E_x$, the score coordinates are
$-E_x^\top z$ and the empirical Hessian matrix at SAA is
$\widehat{\mathsf{H}}_N=\|\bar z_N\|I_2$.
Batch-length tuning is performed separately for each model using 300 trajectories in each of the nine settings formed by $N\in\{1000,4000,16000\}$ and $\lambda\in\{0,0.8,0.95\}$. For each model, we choose one common exponent from $\nu\in\{0.55,0.65,0.75\}$ in $b_N=\lfloor N^\nu\rfloor$ by minimizing the average, over these settings, of the absolute difference between empirical coverage and 0.95 using the $\chi^2_2$ cutoff.



On the sphere the long-run score covariance is
\[
 \mathsf{S}=\tau_\lambda E_{x^*}^\top
 \operatorname{diag}(0.36(1-0.15^2),0.16(1-0.15^2),0)E_{x^*},
\]
used by the oracle benchmark. The feasible Wald statistic uses the exact sphere
logarithm in the estimated tangent space and its estimated covariance.



To compare batch covariances at the empirical and population minimizers, we transport the fitted scores on the sphere using
$\cP_{x\to y}v=v-\langle v,y\rangle(x+y)/(1+\langle x,y\rangle)$.



\subsection{Numerical implementation and regression diagnostics}
\label{sec:rate-diagnostics}



For the sphere instance of the construction in Appendix~\ref{sec:general_manifold_minimax}, use the radial transition function
$S(t)=0$ for $t\le0$, $S(t)=1$ for $t\ge1$, and
\[
 S(t)=\frac{e^{-1/t}}{e^{-1/t}+e^{-1/(1-t)}},\qquad 0<t<1.
\]
In \eqref{eq:chart_potential}, choose
$\eta'(s)=S((s-1/16)/(15/16))$ and $\eta(0)=0$.
The cutoff is $\chi(\Exp_{x_0}u)=1-S((\|u\|-r)/r)$ on the chart.
For centered signs, the full-sphere empirical deviation reduces to
\[
 \Delta_N=r\|\overline Z_N\|
 \max_{0\le t\le2r}t\{1-S((t-r)/r)\}.
\]
We evaluate the maximum on 20,001 equally spaced radial points and solve the chart estimator's monotone scalar response equation by 80 bisection steps.
For the uniform-deviation experiment, a joint regression of the mean log
deviation uses $\log N$, $\log m$, $\log\tau_\lambda$, and
$\log\log N$ over the 73 settings with $N/\tau_\lambda\ge20m$.
The logarithmic regression uses all positive deviations (72985 of 73000
trajectories), and arithmetic means use all trajectories.
The coefficients are $-0.491$, $0.601$, $0.502$, and $-0.086$, with
trajectory-bootstrap intervals $[-0.520,-0.460]$, $[0.597,0.604]$,
$[0.499,0.506]$, and $[-0.320,0.141]$. Regressing the log arithmetic mean
gives $-0.503$, $0.543$, $0.500$, and $0.019$.
The correlation of 0.996 between $\log N$ and $\log\log N$ limits how well this grid separates their effects.


For the modified loss \eqref{eq:simulation-asymmetric-loss}, coordinatewise
minimization brackets the minimum on 4096 equally spaced angles in
$[-\pi,\pi)$, refines it to angular tolerance $10^{-13}$, and applies
Newton refinement to a coordinate gradient or step tolerance $10^{-14}$.
Bisection selects the perturbation size to match the prescribed objective gap
or gradient residual within distance 0.75 of that solution.


\subsection{Batch-exponent tuning and large-sample coverage extension}
\label{sec:exploratory-coverage}
In a separate exploratory study on the torus, all three exponents use the $\chi^2_2$ cutoff throughout.
On the 300 tuning trajectories per setting, the mean absolute coverage errors across the nine sample-size and dependence settings are 0.053, 0.055, and 0.133 for $\nu=0.55, 0.65,$ and $0.75$; on the full 3000-trajectory exploratory ensemble containing those tuning paths, the corresponding values are 0.050, 0.061, and 0.145.
At $N=1000,4000,16000$, $\nu=0.55$ yields 22, 42, and 78 complete batches, respectively, compared with 11, 18, and 29 for $\nu=0.65$.



\begin{figure}[t]
\ifshowrevisions\color{blue}\fi
\centering
\includegraphics[width=\textwidth]{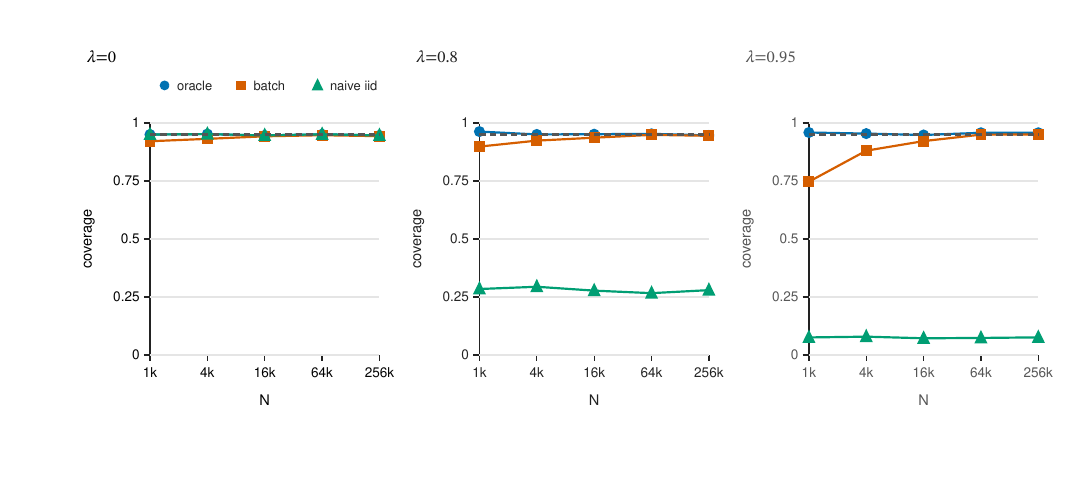}
\caption{Torus coverage across trajectory lengths for $\nu=0.55$. The $N\in\{1000,4000,16000\}$ points show the 3000-trajectory exploratory ensemble, and the $N\in\{64000,256000\}$ extension uses 2000 independent trajectories from a separate sampling stream.}
\label{fig:supp-inference}
\end{figure}



At $N=256000$, feasible coverage in Figure~\ref{fig:supp-inference} ranges from 0.943 to 0.950 across the three dependence levels.


\subsection{Bayes comparison, subspace estimation, and basin localization}

For the torus loss, we enumerate all $2^m$ sign hypotheses and form the
Bayes action under their uniform prior, using the exact refresh-chain
likelihood and known $a,\delta_N,\lambda$. The action averages the population minimizers' angular coordinates
under the posterior; their range lies within a single semicircle.
The design crosses $m\in\{4,8,10\}$, $N\in\{64,256,1024\}$, and
$\lambda\in\{0,0.8,0.95\}$, with $a=1$ and 500 independent trajectories
per setting. The ratio of mean SAA risk to mean Bayes risk ranges from 16.5
to 31.9 and averages 23.7. This comparison quantifies the finite-sample efficiency gain from shrinkage when the exact parametric likelihood and prior are known.

\begin{figure}[t]
\centering
\includegraphics[width=\textwidth]{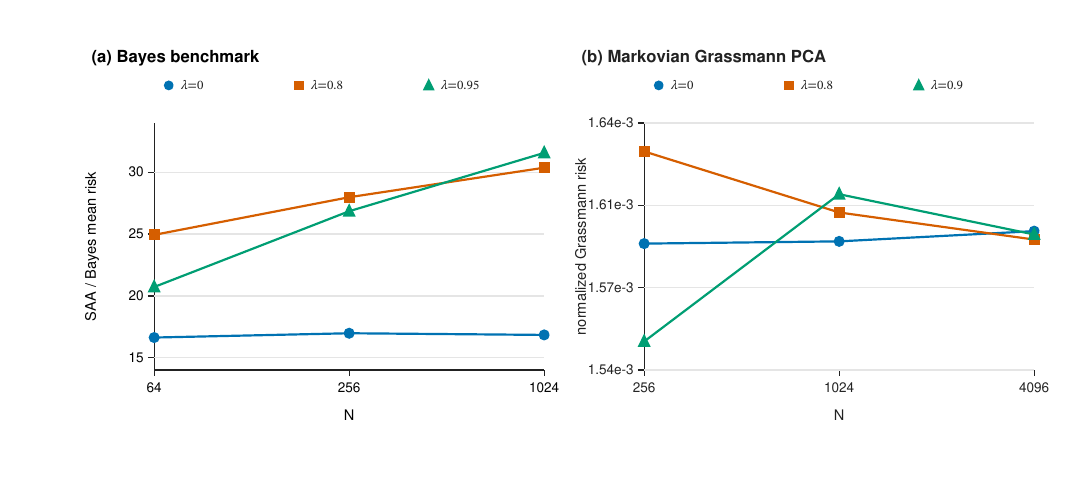}
\caption{Additional parameter-risk comparisons. (a) SAA-to-Bayes mean-risk
ratio on the torus family, averaged over $m\in\{4,8,10\}$.
(b) Dependence-normalized Grassmann risk for ambient dimension $p=50$,
subspace dimension $q=3$, and eigengap 0.1.}
\label{fig:supp-applications}
\end{figure}


For the Grassmann criterion \eqref{eq:grassmann_loss}, use ambient dimension $p$, subspace dimension $q$, and eigengap $h$. Set $\alpha_j=(1-qh)/p+h$ for $j\le q$ and $\alpha_j=(1-qh)/p$ otherwise, and generate bounded observations $z_i=\operatorname{diag}(\sqrt{\alpha_1},\ldots,\sqrt{\alpha_p})Z_i$ from centered whole-vector refresh signs. The population target is the span of the first $q$ coordinate vectors; we measure squared geodesic error.
The design uses $(p,q)\in\{(20,1),(50,3)\}$,
$h\in\{0.1,0.25\}$, $\lambda\in\{0,0.8,0.9\}$, and
$N\in\{256,1024,4096\}$, with 500 trajectories per setting.
The twelve fitted risk slopes have median $-0.990$ and range from $-1.009$ to $-0.920$. Figure~\ref{fig:supp-applications}b shows the dependence-normalized risks for $(p,q)=(50,3)$ and $h=0.1$.


To illustrate the role of basin localization in Theorem~\ref{thm:local_fast_rate}, we use the circle loss
$1-\cos\theta+0.4(1-\cos2\theta)-0.1z\sin\theta$,
centered signs, $N=8192$, and $\lambda=0.8$.
For each of ten independent trajectories, we initialize gradient descent at
361 equally spaced angles in $[-\pi,\pi)$. Armijo backtracking starts
with step 0.5, halves the step, and uses decrease coefficient $10^{-4}$;
near a stationary point with positive second derivative of the empirical loss a Newton refinement is applied.
The stopping tolerance is $10^{-10}$, with at most 10000 iterations.
On each trajectory, 103 initial angles converge to a secondary local stationary point at least
one radian from the global empirical minimizer while satisfying the
gradient tolerance, corresponding to 28.5\% of the deterministic 361-point angular initialization grid.


\FloatBarrier
\clearpage


\end{document}